\pdfoutput=1
\documentclass[10pt,twocolumn]{article}

\usepackage[letterpaper,top=0.85in,bottom=0.9in,left=0.7in,right=0.7in]{geometry}
\usepackage{times}
\usepackage{amsmath,amssymb,amsthm}
\usepackage{graphicx}
\usepackage{booktabs}
\usepackage{multirow}
\usepackage{hyperref}
\usepackage{microtype}
\usepackage{enumitem}
\usepackage{xcolor}
\usepackage{caption}
\usepackage[ruled,vlined,linesnumbered]{algorithm2e}
\SetKwInput{KwIn}{Input}
\SetKwInput{KwOut}{Output}

\hypersetup{pdftitle={Pessimistic-Behavior / Greedy-Target Q-Learning against Adversarial State Perturbations},pdfauthor={Adithya Venkatesh, Xiaolin Sun, Zizhan Zheng},colorlinks=true,linkcolor=blue!50!black,citecolor=blue!50!black,urlcolor=blue!50!black}
\setlist[itemize]{leftmargin=1.1em,itemsep=2pt,topsep=3pt}
\setlist[enumerate]{leftmargin=1.3em,itemsep=2pt,topsep=3pt}
\title{\vspace{-1.0em}Pessimistic-Behavior / Greedy-Target Q-Learning\\against Adversarial State Perturbations\vspace{-0.3em}}
\author{%
Adithya Venkatesh\\
{\small \texttt{adithyakv30@gmail.com}}
\and
Xiaolin Sun\\
{\small Department of Computer Science}\\
{\small Tulane University}\\
{\small New Orleans, LA 70118}\\
{\small \texttt{xsun12@tulane.edu}}
\and
Zizhan Zheng\\
{\small Department of Computer Science}\\
{\small Tulane University}\\
{\small New Orleans, LA 70118}\\
{\small \texttt{zzheng3@tulane.edu}}%
}
\date{}

\begin{document}
\maketitle
\thispagestyle{empty}

\begin{abstract}
Reinforcement learning agents can fail when an attacker perturbs their observations, even if the environment itself is unchanged.
Belief-enriched pessimistic Q-learning defends against such attacks by tracking the set of states consistent with the agent's history and choosing the action whose worst-case value over that set is highest.
It applies this worst-case reasoning in two places: in the Bellman target used to learn the Q-function, and in the rule used to choose actions at test time.
Because both change at once, it is unclear which one provides the robustness, and the test-time search adds computation to every decision.
We separate the two.
We introduce PB-GT (pessimistic-behavior, greedy-target) Q-learning, which collects data with the pessimistic rule but learns and acts greedily, and we evaluate each learned Q-function under both test rules.
On a tabular GridWorld, a continuous GridWorld, and CartPole, most of the robustness comes from the worst-case rule at test time.
The pessimistic Bellman target adds a smaller gain on the tabular GridWorld, and training on adversarial data with a greedy target performs about like standard Q-learning.
We also find that the worst-case rule can be overly cautious when no attacker is present, and we discuss when its extra computation is worth paying.
\end{abstract}

\section{Introduction}
\label{sec:intro}

Reinforcement learning agents trained with Q-learning act by choosing the action with the highest learned value for the state they observe.
This makes them vulnerable to adversarial state perturbations: an attacker who cannot change the environment, but can slightly alter what the agent observes, can make a poor action look best~\cite{huang2017adversarial,zhang2020robust}.
The environment still evolves from the true state; only the agent's input changes.
A value function that is accurate on clean states gives no guarantee about the action it selects on a nearby, perturbed input, and a small perturbation at a critical state can change the outcome of an episode.
Our discussion concerns Q-learning and other methods that act greedily with respect to a learned value function; policy-based methods choose actions differently.

Existing defenses take several approaches.
Regularization methods such as SA-DQN~\cite{zhang2020robust} and WocaR-RL~\cite{liang2022wocar} train the policy to keep its action unchanged under small perturbations, and they remain cheap at test time because the deployed policy is a single forward pass.
ATLA~\cite{zhang2021robust} instead trains a learned attacker alongside the defender.
This can produce a strong attacker and a defender that is robust in practice, but the alternating two-player training is expensive and can be unstable.
Sun and Zheng~\cite{sun2024belief} take a third approach, belief-enriched pessimistic Q-learning.
The agent keeps track of the set of true states that are consistent with its observations and actions so far, and it chooses the action whose worst-case value over that set is highest.
It uses the same worst-case reasoning when computing the Bellman target during training.

This method therefore applies worst-case reasoning in two places: when learning the Q-function and when choosing actions at test time.
The test-time step has a cost.
Every decision requires updating the set of possible states and searching it for the worst case, while a standard Q-learning agent only compares one value per action.
This raises the question we study: is the robustness of belief-enriched pessimistic Q-learning a property of the learned Q-function, or of the worst-case rule applied at test time?
If training with worst-case reasoning already makes the Q-function robust, an agent could be trained pessimistically and deployed with the cheaper greedy rule.
If the robustness comes from the test-time rule, that cost cannot be avoided this way.

To separate the two, we introduce PB-GT (pessimistic-behavior, greedy-target) Q-learning.
PB-GT collects data the same way as pessimistic Q-learning, under attack and with the worst-case action rule, but it learns with the standard greedy Bellman target and acts greedily at test time.
Because the two methods learn different Q-functions, they also visit different states during training, so comparing them directly mixes several effects.
We therefore evaluate each learned Q-function under both test rules, which gives four variants: pessimistic Q-learning as proposed (pessimistic $Q$-function, worst-case test rule), PB-GT as proposed (PB-GT $Q$-function, greedy test rule), and two swapped variants, the pessimistic $Q$-function with a greedy test rule and the PB-GT $Q$-function with the worst-case test rule (Table~\ref{tab:swap}).
Comparing the two test rules on the same $Q$-function isolates the effect of the test rule.
Comparing the pessimistic and PB-GT $Q$-functions under the same worst-case test rule, in particular the PB-GT $Q$-function with the worst-case test rule against pessimistic Q-learning as proposed, shows what the pessimistic Bellman target adds on top of that rule, together with the different states it leads the agent to visit during training.

We run this comparison on a tabular GridWorld (\S\ref{sec:gw-setup}), a continuous version of the same grid (\S\ref{sec:cgw-setup}), and CartPole (\S\ref{sec:cp-setup}).
In each setting, most of the protection under attack comes from the worst-case test rule.
The pessimistic Bellman target adds a smaller further gain on the tabular GridWorld and no measurable gain on the continuous GridWorld, and PB-GT performs about like standard Q-learning under attack.
We also find a cost beyond computation: on the tabular GridWorld, the worst-case rule is overly cautious when no attacker is present, because the set of possible states rarely narrows to a single state.

\paragraph{Contributions.}
\begin{enumerate}[leftmargin=1.4em]
\item We implement standard Q-learning and belief-enriched pessimistic Q-learning, following Algorithms~1--2 of~\cite{sun2024belief}, on a tabular GridWorld and evaluate them over $500$ random seeds against a large set of learned attackers, including attackers that observe the agent's belief.
\item We define PB-GT, which keeps the pessimistic data collection but uses the standard greedy Bellman target and test rule.
\item We evaluate four variants, each learned Q-function under both test rules, including the PB-GT $Q$-function with the worst-case test rule, to measure how much of the robustness comes from the test rule and how much from the pessimistic Bellman target.
\item On a continuous two-dimensional version of the same grid and on CartPole-v1, we replace a learned particle filter with a set that is guaranteed to contain the true state, and we report the same comparison.
\item Code, the final CSV files, and the figure script are listed in Appendix~\ref{app:code}.
\end{enumerate}

\section{Problem formulation}
\label{sec:bg}

\subsection{MDPs and Q-learning}
A Markov decision process (MDP) is a tuple $\langle \mathcal{S},\mathcal{A},P,R,\gamma\rangle$.
The set $\mathcal{S}$ is the state space and $\mathcal{A}$ is a finite action space.
The kernel $P(s'\mid s,a)$ is the probability of moving to $s'$ after action $a$ in state $s$.
The function $R(s,a)$ is the immediate reward.
The scalar $\gamma\in(0,1]$ is the discount factor.
The tabular GridWorld in this paper is a finite MDP with $\gamma=1$.
The continuous GridWorld and CartPole have continuous states, finite actions, and $\gamma<1$.

At time $t$ the environment occupies a true state $s_t\in\mathcal{S}$.
The agent selects $a_t\in\mathcal{A}$, receives $R(s_t,a_t)$, and the environment moves to $s_{t+1}\sim P(\cdot\mid s_t,a_t)$.
A deterministic policy $\pi$ maps states, or observations, to actions.
The optimal action-value function satisfies
\begin{equation}
Q^\star(s,a)=R(s,a)+\gamma\sum_{s'}P(s'\mid s,a)\max_{a'}Q^\star(s',a').
\label{eq:bellman}
\end{equation}
When $Q^\star$ is known, the optimal policy is $\pi^\star(s)=\arg\max_a Q^\star(s,a)$.

Vanilla Q-learning~\cite{watkins1992q} estimates $Q^\star$ from sampled transitions.
After $(s,a,R,s')$ it applies
\begin{equation}
Q(s,a)\leftarrow (1-\alpha)Q(s,a)+\alpha\bigl(R+\gamma\max_{a'}Q(s',a')\bigr),
\label{eq:vanilla-q}
\end{equation}
where $\alpha\in(0,1]$ is a learning rate.
The expression in parentheses is the \emph{Bellman target}, also called the TD target: the sampled reward plus a discounted value at the next state.
This paper uses ``Bellman target'' for that update value.
During data collection the agent is $\eta$-greedy on the true state: with probability $1-\eta$ it takes $\arg\max_a Q(s,a)$, and with probability $\eta$ it takes a uniform random action.
We write $\eta$ for this exploration rate to keep it separate from the attack budget $\varepsilon$ introduced in \S\ref{sec:threat}.
Deep Q-networks~\cite{mnih2015human} replace the table with a network $Q_\theta$ trained from a replay buffer.
In the deep experiments we use the Double DQN target~\cite{van2016deep}, which takes the greedy action of the online network and scores it with the target network, so that the Bellman target does not use the same $\max$ that selected the action.
The tabular experiments do not use Double DQN.
In all of these methods, the deployed policy is $\arg\max_a Q(\cdot,a)$ on whatever input the agent is handed.

\subsection{Adversarial state perturbations}
\label{sec:threat}

We imagine an attacker during adversarial training to obtain a robust policy, while the actual attacker only appears at test time.
In the experiments, fine-tuning therefore trains against an attacker we control, and evaluation measures return both with and without an attacker.
The initial vanilla pretraining does not use an attacker.
When an attacker is present, it changes only the observation.
The environment still occupies $s_t$ and still transitions from $s_t$~\cite{huang2017adversarial,zhang2020robust,sun2024belief}.

At step $t$ an attacker who sees $s_t$ produces
\begin{equation}
\tilde s_t \in B_\varepsilon(s_t).
\label{eq:ball}
\end{equation}
In continuous domains $B_\varepsilon(s)$ is the $\ell_\infty$ ball of radius $\varepsilon$ centered at $s$, intersected with the state domain.
On the tabular GridWorld, $B_\varepsilon(s)$ is the discrete one-step neighborhood defined in \S\ref{sec:gw-setup}.
The agent sees only $\tilde s_t$ and must choose $a_t$ from that observation.
The environment then transitions from the true state, $s_{t+1}\sim P(\cdot\mid s_t,a_t)$, and emits the true reward $R(s_t,a_t)$.
The next observation, if an attacker is present, is a perturbation of $s_{t+1}$.

Three properties of this model matter for a defense.
The attacker is observation-only, so any loss comes from the agent choosing a poor action, not from a change to dynamics or rewards.
The attacker is better informed than the defender: it knows $s_t$, and the strongest attacks we use also know the defender's $Q$.
The defender sees only $\tilde s_t$.
The interaction is a Stackelberg game with asymmetric observations.
For a fixed defender $\pi$, the attacker's problem is itself an MDP~\cite{zhang2021robust,sun2024belief}.
The attacker's state is $s_t$, its action is $\tilde s_t\in B_\varepsilon(s_t)$, and its reward is $-R(s_t,\pi(\tilde s_t))$.
We use that fact when we freeze a defender and train a DQN to attack it.

While learning, the agent knows $\varepsilon$, or an upper bound on it.
In both the tabular and the deep experiments, the Bellman target is computed from the true state $s_t$ and the true next state $s_{t+1}$, while the behavior policy acts on the attacked observation.
This is a modeling choice about what the training update may use; at deployment the agent sees only the observation.

Semantic perturbations that leave the $\varepsilon$-ball, including the diffusion attacks of~\cite{sun2025shift}, sit outside this threat model.
Every attack reported below is required to land in $B_\varepsilon(s)$.

\subsection{Notation}
Table~\ref{tab:notation} fixes the notation used below.
We write $\bar s$ and $m$ for candidate true states in an uncertainty set, $\tilde s$ for the perturbed observation, and $s$ for the actual environment state.
Below, \emph{pessimistic $Q$} means the $Q$-function learned by pessimistic Q-learning, and \emph{PB-GT $Q$} means the $Q$-function learned by PB-GT.

\begin{table}[t]
\centering
\small
\caption{Notation used throughout the paper.}
\label{tab:notation}
\begin{tabular}{@{}ll@{}}
\toprule
Symbol & Meaning \\
\midrule
$s_t\in\mathcal{S}$ & true environment state at step $t$ \\
$\tilde s_t$ & perturbed observation given to the agent \\
$\bar s,\;m$ & candidate true state in an uncertainty set \\
$\mathcal{A},\,a_t$ & action set; action executed at step $t$ \\
$B_\varepsilon(s)$ & admissible perturbation set \\
$M_t\subseteq\mathcal{S}$ & belief: states consistent with history \\
$\omega_\pi$ & attacker's best response to defender $\pi$ \\
$Q,\,Q_\theta$ & tabular or parameterized action values \\
$\gamma,\,\alpha$ & discount factor; learning rate \\
$\varepsilon$ & attack budget (radius of $B_\varepsilon$) \\
$\eta$ & exploration rate ($\eta$-greedy) \\
$n$ & number of independent seeds \\
\bottomrule
\end{tabular}
\end{table}

\section{Belief-enriched pessimistic Q-learning and PB-GT}
\label{sec:pes}

PB-GT is a modification of the method of~\cite{sun2024belief}, so we first write the version we implement.
The inner operator is a hard minimum over the belief.

\subsection{The maximin policy}
For each action the agent reads $Q$ at every candidate true state that could have produced the observation, keeps the smallest value, and then chooses the action whose worst-case value is largest:
\begin{equation}
\pi(\tilde s)=\arg\max_{a\in\mathcal{A}}\;\min_{\bar s\in B_\varepsilon(\tilde s)} Q(\bar s,a).
\label{eq:maximin}
\end{equation}
An average over the ball would mix a bad candidate with harmless ones.
The attacker is free to pick the bad candidate, and it sees more than the defender does, so the hard $\min$ is the operator that answers the attack we are studying.
In the tabular GridWorld the inner minimum is exact: it evaluates $Q$ at every grid state still in the belief.
In the continuous GridWorld and on CartPole the belief is a box, and the inner minimum is a hard minimum over a lattice of points on that box.
The continuous grid uses a $3{\times}3$ lattice.
CartPole uses $16$ points, two along each coordinate of the box.
A finer lattice costs more.
The wall-clock factors in \S\ref{sec:speed} are the cost of these searches, not a fixed property of the maximin rule.
We use Eq.~\eqref{eq:maximin} as the pessimistic action rule.
Pessimistic Q-learning also places the same $\min$ inside the Bellman target below.
PB-GT does not.

Against such a defender, the attacker's best response under the white-box objective we use in training is
\begin{equation}
\omega_\pi(s)=\arg\min_{\tilde s\in B_\varepsilon(s)} Q\bigl(s,\pi(\tilde s)\bigr).
\label{eq:omega}
\end{equation}
The attacker searches the ball for the observation that induces the action with the lowest true value.
We use Eq.~\eqref{eq:omega} to generate adversarial observations during tabular fine-tuning.
At evaluation we replace it with the learned attackers of \S\ref{sec:setup}.

The pessimistic Bellman target replaces the greedy target of Eq.~\eqref{eq:vanilla-q} with a maximin target over the propagated belief intersected with the $\varepsilon$-ball of the true next state:
\begin{equation}
y = R + \gamma \max_{a'} \min_{m \in M' \cap B_\varepsilon(s')} Q(m, a').
\label{eq:pes-target}
\end{equation}
Here $M'$ is the belief after the current step.
The target uses the true next state $s'$, in the tabular setting and in the deep settings, which matches the training assumption in \S\ref{sec:threat}.
Pessimistic Q-learning uses the maximin rule both to choose the training action and inside this Bellman target.
Whether this update converges is discussed in \S\ref{sec:pbgt}.

\subsection{Belief tracking}
\label{sec:belief}

Taking the worst case over the entire ball $B_\varepsilon(\tilde s)$ contains the true state, and it is also cautious.
Many members of the ball are inconsistent with the actions the agent has already taken.
The belief $M_t\subseteq\mathcal{S}$ is the set of true states still consistent with the history.
We call a belief \emph{sound} when the set is guaranteed to contain the true state.
Restricting the $\min$ to $M_t$ tightens the worst case and, when the update is exact, keeps the belief sound.

Our tabular implementation maintains $M_t$ in two stages, matching the tabular algorithm of~\cite{sun2024belief}.
Given $M_t$ and the observation $\tilde s_t$, the agent filters:
\begin{equation}
M'_t = M_t \cap B_\varepsilon(\tilde s_t).
\label{eq:belief-filter}
\end{equation}
After action $a_t$ it propagates the survivors through the dynamics:
\begin{equation}
M_{t+1} = \bigl\{\,s' : \exists\, m\in M'_t,\; P(s'\mid m,a_t)>0 \,\bigr\}.
\label{eq:belief}
\end{equation}
The belief starts at the full state set, $M_0=\mathcal{S}$, and we reset it at the start of every episode.
If $s_t\in M_t$ and the neighborhood is symmetric, then $s_t\in M'_t$, because $\tilde s_t\in B_\varepsilon(s_t)$ implies $s_t\in B_\varepsilon(\tilde s_t)$.
Combined with $M_0=\mathcal{S}$, the true state is never discarded by a correct update.
On the deterministic grid the observation repeatedly intersects a small ball, so $|M_t|$ usually shrinks.
The policy under attack is
\begin{equation}
\pi(\tilde s)=\arg\max_{a\in\mathcal{A}}\;\min_{m\in M \cap B_\varepsilon(\tilde s)} Q(m,a).
\label{eq:belief-policy}
\end{equation}
When the intersection is a singleton, Eq.~\eqref{eq:belief-policy} reduces to a greedy $\arg\max$ at that state.
If the intersection is empty we fall back to a greedy $\arg\max$ at the observed state.
In the belief log of \S\ref{sec:gw} that event did not occur.

\subsection{Three design axes}
\label{sec:methods}

The methods compared in the paper share an environment and an attacker protocol.
In the deep case they also share a vanilla backbone and the same pretraining budget.
They differ along three axes: the observations and the behavior rule used while collecting data, the Bellman target, and the test rule.
Vanilla Q-learning is greedy on all three.
Belief-enriched pessimistic Q-learning is pessimistic on all three.
PB-GT is pessimistic on the first axis and greedy on the other two.
Table~\ref{tab:methods} records the three training procedures.
Table~\ref{tab:swap} records the four evaluation configurations used to separate the test rule from the learned $Q$.
The two off-diagonal configurations are the extra rows in the later tables.
Each one keeps a trained $Q$-function and changes only the test rule.

\begin{table}[t]
\centering
\small
\caption{The three training procedures, along the axes of \S\ref{sec:methods}.}
\label{tab:methods}
\setlength{\tabcolsep}{3pt}
\begin{tabular}{@{}lccc@{}}
\toprule
 & Vanilla & Pessimistic & PB-GT \\
\midrule
Train behavior & greedy on $s$ & maximin on $\tilde s$ & same as pes. \\
TD target & greedy on $s'$ & maximin over $M'$ & greedy on $s'$ \\
Test & greedy on $\tilde s$ & maximin over $M$ & greedy on $\tilde s$ \\
Belief at test & no & yes & no \\
\bottomrule
\end{tabular}
\end{table}

\begin{table}[t]
\centering
\small
\caption{The four evaluation configurations ($Q$-function $\times$ test rule).
Off-diagonal rows are the swap: the trained $Q$ stays, and only the test rule changes.
``Pessimistic $Q$'' is the $Q$-function learned by pessimistic Q-learning.
``PB-GT $Q$'' is the $Q$-function learned by PB-GT.}
\label{tab:swap}
\setlength{\tabcolsep}{4pt}
\begin{tabular}{@{}lll@{}}
\toprule
Learned $Q$ & Test rule & Role \\
\midrule
Pessimistic $Q$ & maximin & trained rule \\
Pessimistic $Q$ & greedy & test rule swapped \\
PB-GT $Q$ & greedy & trained rule \\
PB-GT $Q$ & maximin & test rule swapped \\
\bottomrule
\end{tabular}
\end{table}

\subsection{Pessimistic Q and PB-GT}
\label{sec:pbgt}

Algorithm~\ref{alg:pes-tab} is the tabular pessimistic loop we run.
We start from a copy of a pretrained vanilla $Q$, reset the belief to $\mathcal{S}$ at each episode, form $\tilde s=\omega_\pi(s)$, take an $\eta$-greedy maximin action, and update $Q(s,a)$ with Eq.~\eqref{eq:pes-target}.
At deployment the agent updates $M_t$ and acts by Eq.~\eqref{eq:belief-policy}, including on episodes with no attacker.
The clean return we report for this method is the return of that deployed maximin rule, including when there is no attacker.
A greedy readout of the same $Q$-function, $a=\arg\max_a Q(s,a)$ on the true state, is reported separately so that a cautious maximin can be distinguished from a $Q$-function that cannot reach the goal.
Sun and Zheng~\cite{sun2024belief} evaluate natural reward with the deployed belief and maximin rule.
We do the same.

PB-GT uses the data-collection rule of pessimistic Q-learning and changes the Bellman target and the test rule.
Algorithm~\ref{alg:pbgt-tab} is the tabular loop.
The step has the same shape as Algorithm~\ref{alg:pes-tab}: an adversarial observation, a belief, and a maximin behavior rule.
The learned $Q$ is not the same function.
The two updates write different targets, so the action taken on the next step can differ, and the distribution of visited pairs $(s,a)$ can differ with it.
The update is the ordinary greedy Bellman target of Eq.~\eqref{eq:vanilla-q}, written again as
\begin{equation}
Q(s,a)\leftarrow (1-\alpha)Q(s,a)+\alpha\bigl(R+\gamma\max_{a'}Q(s',a')\bigr).
\label{eq:pbgt-tab}
\end{equation}
In the tabular setting this is a plain Q-learning target.
There is no second network and no Double DQN correction.
The belief decides which action rule is used during training.
It is not used in the target.
At deployment the belief and the maximin are dropped,
\begin{equation}
a=\arg\max_{a} Q(\tilde s,a),
\label{eq:pbgt-test}
\end{equation}
so the inference cost matches vanilla: one table lookup, or one forward pass in the deep case.

Substituting Eq.~\eqref{eq:pbgt-tab} for Eq.~\eqref{eq:pes-target} changes the operator being learned.
Pessimistic Q-learning applies a maximin Bellman operator.
Sun and Zheng~\cite{sun2024belief} show that this operator is not a contraction in general, so a fixed point need not exist.
We do not claim that its learned values are the value of the maximin policy.
PB-GT applies the standard optimality operator.
When the discount is strictly less than one, that operator is a contraction, and with enough coverage of the state-action space the update converges to $Q^\star$ of the clean MDP.
The continuous GridWorld and CartPole use such a discount.
The tabular GridWorld uses $\gamma=1$, so the same target is still the optimality operator of the clean MDP, but the contraction argument does not apply there.
Pessimistic behavior changes which states are visited.
It does not, by itself, change the operator.
What it can change is coverage: a state the agent never visits is not updated.
A greedy policy read from an accurate $Q^\star$ can still be exploited, because the attacker chooses $\tilde s$ so that $\arg\max_a Q^\star(\tilde s,a)$ is a poor action at the true $s$.
The swap in Table~\ref{tab:swap} takes each trained $Q$-function and changes only the test rule.
A gap between the two test rules on one $Q$-function is a gap between those rules.
A gap between pessimistic $Q$ and PB-GT $Q$ under one test rule mixes the Bellman target with the different states visited in training.

\begin{algorithm}[t]
\caption{Belief-enriched pessimistic Q (tabular)}
\label{alg:pes-tab}
\KwIn{pretrained $Q$; budget $\varepsilon$; episodes $E$}
\KwOut{fine-tuned $Q$}
$Q\leftarrow$ copy of pretrained table\;
\For{$e=1$ \KwTo $E$}{
  reset $s$; $M\leftarrow\mathcal{S}$\;
  \While{$s$ not terminal and step cap not hit}{
    $\pi(\cdot)\leftarrow\arg\max_a\min_{m\in M\cap B_\varepsilon(\cdot)}Q(m,a)$\;
    $\tilde s\leftarrow\omega_\pi(s)$\;
    $a\leftarrow\eta$-greedy w.r.t.\ $\pi(\tilde s)$\;
    execute $a$, observe $R,s'$\;
    $M\leftarrow$ filter and propagate, Eqs.~\eqref{eq:belief-filter}--\eqref{eq:belief}\;
    $y\leftarrow R+\gamma\max_{a'}\min_{m\in M\cap B_\varepsilon(s')}Q(m,a')$\;
    $Q(s,a)\leftarrow (1-\alpha)Q(s,a)+\alpha\,y$\;
    $s\leftarrow s'$\;
  }
}
\textbf{Test:} act by Eq.~\eqref{eq:belief-policy}, with or without an attacker\;
\end{algorithm}

\begin{algorithm}[t]
\caption{PB-GT Q-learning (tabular)}
\label{alg:pbgt-tab}
\KwIn{pretrained $Q$; budget $\varepsilon$; episodes $E$}
\KwOut{fine-tuned $Q$}
$Q\leftarrow$ copy of pretrained table\;
\For{$e=1$ \KwTo $E$}{
  reset $s$; $M\leftarrow\mathcal{S}$\;
  \While{$s$ not terminal and step cap not hit}{
    $\pi(\cdot)\leftarrow\arg\max_a\min_{m\in M\cap B_\varepsilon(\cdot)}Q(m,a)$\;
    $\tilde s\leftarrow\omega_\pi(s)$\;
    $a\leftarrow\eta$-greedy w.r.t.\ $\pi(\tilde s)$\;
    execute $a$, observe $R,s'$\;
    \textbf{greedy Bellman target:}
    $Q(s,a)\leftarrow (1-\alpha)Q(s,a)+\alpha(R+\gamma\max_{a'}Q(s',a'))$\;
    $M\leftarrow$ filter and propagate\;
    $s\leftarrow s'$\;
  }
}
\textbf{Test:} $a=\arg\max_a Q(\tilde s,a)$; no belief and no maximin\;
\end{algorithm}

\subsection{Deep settings and the sound box}
\label{sec:deep}

This is not a reproduction of BP-DQN.
We do not train the particle-filter RNN of~\cite{sun2024belief}, and we do not run their $10{\times}10$ maze or Atari suite.
The deep experiments use a sound set-membership box in place of a learned belief.
Sound is the property from \S\ref{sec:belief}: the set is guaranteed to contain the true state.
On the continuous GridWorld and on CartPole that box uses the true dynamics model.
BP-DQN's learned belief does not need that model.
Appendix~\ref{app:box} writes the box update in one place.
A failed GRU-particle port of the CartPole agent is kept in Appendix~\ref{app:cp-extra} as a record of that attempt, and it is not a column of the main comparison.

In the deep case, pessimistic and PB-GT fine-tune from a pretrained vanilla DQN.
On the continuous GridWorld the backbone is not frozen.
On the trained CartPole run we freeze it and train only the last linear layer.
Data collection uses the pessimistic behavior rule: an attacker produces $\tilde s\in B_\varepsilon(s)$, and the action is $\eta$-greedy maximin over the current box.
PB-GT then forms Double DQN greedy targets on the true next state.
The pessimistic Bellman target is a maximin over the propagated next box.
Double DQN is used only in these deep settings.
Testing for PB-GT is a greedy $\arg\max$ on $\tilde s$.
Testing for the pessimistic agent is maximin over the box, including on clean episodes.
Pseudocode for the deep PB-GT loop is Appendix~\ref{app:alg}.

\section{Experimental setup}
\label{sec:setup}

Returns below are means across seeds, with the sample standard deviation (ddof${=}1$), unless a sentence says otherwise.
Win rates count a seed only when one method's return is strictly larger.

\subsection{Tabular GridWorld}
\label{sec:gw-setup}

The discrete testbed is a $5{\times}5$ grid.
Every non-terminal transition costs $-1$.
A gold cell at $(0,3)$ yields $+10$ and ends the episode.
A bomb cell at $(4,3)$ yields $-10$ and ends the episode.
The four move actions are deterministic, and a move into a wall leaves the agent in place.
We set $\gamma=1$ and anneal the learning rate.
On this grid the admissible set is the discrete one-step neighborhood
\begin{equation}
B_\varepsilon(s)=\{s\}\cup\{\text{on-grid cells adjacent to }s\},
\label{eq:disc-ball}
\end{equation}
so $|B_\varepsilon(s)|\le 5$.
Episodes are capped at $1000$ steps, so an attacker can hurt the agent by steering it into the bomb or by trapping it in a loop that only accumulates step costs.

Each seed starts with $2000$ episodes of vanilla Q-learning.
Pessimistic Q-learning and PB-GT then fine-tune for $500$ episodes from that same table.
A configuration here is one learned $Q$-function under one test rule and one attack condition.
The four $Q$-function $\times$ test-rule rows are Table~\ref{tab:swap}.
We evaluate $100$ episodes in each configuration on each of $500$ seeds, again capped at $1000$ steps.

\paragraph{Learned attackers.}
Evaluation attacks are learned and then frozen.
We train three Stable-Baselines3 DQN attackers for each of five defender rules: vanilla, pessimistic $Q$ with maximin test, PB-GT $Q$ with greedy test, pessimistic $Q$ with greedy test, and PB-GT $Q$ with maximin test.
Together those are fifteen attackers.
Each of the three attackers for a rule is trained against a different defender seed ($99999$, $99998$, or $99997$) and a different attacker seed.
The attacker's action is a direction in $\{\mathrm{UP},\mathrm{DOWN},\mathrm{LEFT},\mathrm{RIGHT},\mathrm{STAY}\}$, applied to the true cell and clipped to the grid, so the observation lies in $B_\varepsilon(s)$.
$\mathrm{STAY}$ reports the true cell, so ``no attack'' is already in every attacker's action set.
The attacker's reward is the negation of the defender's reward.
Those fifteen attackers observe the true state but not the defender's belief.

Against the two maximin rules we also train six \emph{belief-aware} attackers: three for pessimistic $Q$ with maximin test and three for PB-GT $Q$ with maximin test, again against defender seeds $99999$, $99998$, and $99997$, with the same training budget and architecture.
Each belief-aware attacker observes the true cell together with a $25$-dimensional binary mask of the defender's current belief, so its policy can condition on the history-dependent set that the maximin uses.
All five deployed rules are scored against these six attackers as well.

A \emph{matched attacker} is an attacker trained against the same decision rule it is evaluated on.
On this tabular grid the attackers are trained on three defender seeds and then reused, so the match is to the rule, not to every evaluated seed.
\emph{Matched} return for a greedy or vanilla rule is the mean, on that seed, of that rule's own three (belief-blind) attackers.
For a maximin rule, matched return in the main table is the mean of that rule's own three belief-aware attackers.
\emph{Worst-of-22} return is the per-seed minimum over all twenty-one learned attackers (the original fifteen plus the six belief-aware ones) and the clean (no-attack) return under the same deployed rule.
Because $\mathrm{STAY}$ is already an attacker action, the worst case per seed must include no attack.
Appendix~\ref{app:worst910} reports the same $500$-seed evaluation under further summaries: belief-blind matched return for the maximin rows, and worst-of-9, worst-of-10, worst-of-15, and worst-of-16.

A single frozen attacker is a poor summary on this grid.
The sample standard deviation of a method's three own-attacker means (each mean taken across the $500$ defender seeds) is $157.64$ for vanilla, $64.14$ for pessimistic $Q$ with maximin test under the belief-blind own three, $2.80$ for pessimistic $Q$ with greedy test, $146.26$ for PB-GT $Q$ with maximin test under the belief-blind own three, and $102.19$ for PB-GT $Q$ with greedy test.
Those spreads are why every attack column is either a mean over a rule's own three attackers or a minimum over the full set, rather than the return against one attacker.

Inference time is wall-clock milliseconds per decision on a single process.
The timer starts after the attacker has produced $\tilde s$ and stops before the environment step.
For the pessimistic agent the measured path includes the belief update and the maximin.
For vanilla and PB-GT it is the $\arg\max$.
We time seeds $0$--$19$ on the attack path of those three deployed rules.
The swapped configurations in Table~\ref{tab:swap} use the same code paths, so the timing table reuses those measurements.
A separate timing of greedy readouts on clean episodes is not used as the cost of maximin.

\subsection{Continuous GridWorld}
\label{sec:cgw-setup}

\begin{figure}[t]
\centering
\includegraphics[width=\columnwidth]{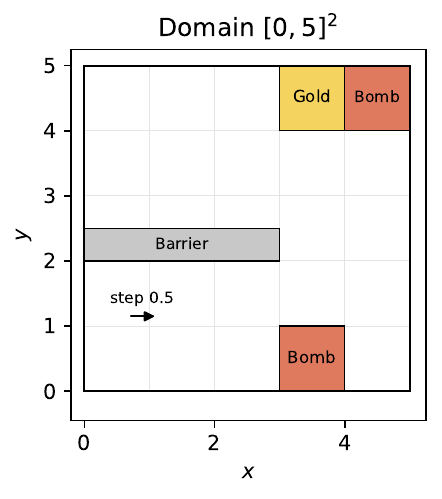}
\caption{The continuous GridWorld of \S\ref{sec:cgw-setup}.
The domain is $[0,5]^2$ and each action moves the agent by $0.5$.
Gold is the image of tabular cell $(0,3)$.
The lower bomb is the image of tabular cell $(4,3)$.
A second bomb touches the gold, and the barrier leaves a gap for $x>3$.}
\label{fig:cgw-map}
\end{figure}

The function-approximation testbed is a continuous-state lift of the same $5{\times}5$ grid.
States are pairs $(x,y)\in[0,5]^2$.
The four actions translate the agent by $0.5$, and a move into a wall leaves the agent in place.
Gold and bomb rectangles correspond to the tabular cells $(0,3)$ and $(4,3)$, with an extra bomb adjacent to the gold and a barrier that forces a corridor.
Figure~\ref{fig:cgw-map} draws the domain, the gold, both bombs, the barrier, and the step size.

Every non-terminal step costs $-1$.
Episodes are capped at $100$ steps.
The $Q$-network is a $2{\to}128{\to}128{\to}4$ MLP trained with Double DQN.
The production comparison uses $\varepsilon{=}0.25$, $80$k vanilla steps, $25$k fine-tuning steps, $50$ evaluation episodes, and $10$ seeds.
The target/test swap on this environment is the four configurations of Table~\ref{tab:swap}: each off-diagonal entry keeps the trained $Q$-function and changes only the test rule.
It was run for seeds $0$--$4$ only, and it is labeled as five seeds wherever it is reported.
We chose $\varepsilon{=}0.25$ from the shorter sweeps in Appendix~\ref{app:eps}, where the pessimistic greedy clean return is least damaged, and the final $80$k+$25$k schedule was run only at $0.25$.

The belief is the sound box of Appendix~\ref{app:box}.
We also evaluate a memoryless sound $\varepsilon$-ball: a $3{\times}3$ lattice on $B_\varepsilon(\tilde s)$ with no temporal intersection.
In both cases the inner operator is a hard minimum over the lattice.
PB-GT uses the box during training and discards it at test time.

Each seed is evaluated on several suites, defined here so the tables can stay short.
A \emph{clean} episode has no attacker.
The agent uses the rule it deploys: maximin over the box for pessimistic methods, and a greedy readout for vanilla and for PB-GT.
We also log a greedy readout of the pessimistic weights on clean states.
A \emph{shared-random} episode adds $\delta\sim\mathrm{Unif}([-\varepsilon,\varepsilon]^2)$ from a shared random stream, so every method sees the same noise sequence.
A \emph{PGD-10} episode runs a $10$-step projected gradient attack~\cite{madry2018towards} with the MinBest objective~\cite{huang2017adversarial,zhang2020robust} on $Q(\tilde s, a^\star)$, where $a^\star$ is the greedy action at the true state.
A \emph{matched search} episode uses an attacker matched, in the sense above, to the decision rule being evaluated: it scores $32$ candidates by $Q(s,\pi(\tilde s))$ under that rule.
A \emph{matched SB3} episode uses a DQN attacker trained for $40$k steps against that defender and seed, with a $5{\times}5$ perturbation grid inside the ball.

\subsection{CartPole-v1}
\label{sec:cp-setup}

The third setting is Gymnasium CartPole-v1.
Observations are scaled to $[-1,1]$ for the network.
The attack budget is $\varepsilon{=}0.05$ in raw coordinates under $L_\infty$.
The vanilla agent is a Double DQN with a $[128,128]$ MLP, trained for $5{\times}10^5$ steps and continued until its average return reaches $475$, CartPole-v1's solved threshold.
We run seeds $0$--$4$ on one CPU thread.

Pessimistic and PB-GT clone that checkpoint, freeze the backbone, and train the last linear layer at learning rate $10^{-5}$.
The belief is a sound filter-then-propagate box, sound in the sense of \S\ref{sec:belief}: the box is guaranteed to contain the true state.
It uses interval dynamics, and the inner min is a hard min over $16$ lattice points.
That lattice is the search from \S\ref{sec:pes}, and its size is part of the cost.
Behavior during fine-tuning is $\eta$-greedy maximin on an attacked observation.
The pessimistic Bellman target is a maximin over the next box, and the deployed test rule is that same maximin, including on clean episodes.
PB-GT uses a greedy Double DQN target on the true next state and a greedy test policy.
The pessimistic budget is capped at $5{\times}10^4$ steps.
Every $5{\times}10^3$ steps, including step~$0$, we record clean return under the deployed maximin rule and keep the weights with the best deployed clean return.
If that return has reached $475$ and a later evaluation falls below $400$, training stops and the best checkpoint is restored.
PB-GT fine-tunes for $4{\times}10^4$ steps.

Each configuration is one method under one attack condition.
Evaluation uses $50$ episodes in each configuration on each of five seeds (seeds $0$--$4$).
A shared-random episode adds $\delta\sim\mathrm{Unif}([-\varepsilon,\varepsilon]^4)$ from a shared stream.
A \emph{PGD*} episode draws $32$ candidates inside $B_\varepsilon(s)$ and scores $Q(s,\pi(\tilde s))$.
PGD* is that random search.
It is not the $10$-step iterative PGD used on the continuous GridWorld.
The unshaped matched attacker is a DQN trained for $40$k steps on the environment return.
A configuration is vacuous when that attacker never ended an episode during its training.
The environment reward is too sparse for the attacker to learn from, because the pole can stay up for a long time before an episode ends.
A \emph{shaped attacker} uses the same budget and the reward $|\theta|+0.1|x|$, which grows with the pole angle and the cart position, so the attacker gets a signal before the episode ends.
Inference time is single-process wall clock per action, summarized as the mean of the five per-seed means.
Pole-sensor attacks, an $\varepsilon$ sweep, a position-sensor variant, the frozen-weight rule swap, and the failed GRU port are Appendix~\ref{app:cp-extra}.

\section{Results}
\label{sec:results}

\subsection{Tabular grid: the test rule carries most of the gap}
\label{sec:gw}

\begin{table*}[t]
\centering
\small
\caption{Tabular $5{\times}5$ GridWorld, $500$ seeds, sample std (ddof${=}1$).
Clean uses the deployed rule and also reports the median.
The greedy-readout column is a diagnostic for the maximin-test rows.
Matched (own~3) averages a rule's own three attackers: belief-aware for the two maximin rows, belief-blind for the greedy and vanilla rows.
Worst-of-22 is the per-seed minimum over all twenty-one learned attackers and the clean return; ``no attack'' (the attacker always reporting the true cell, i.e.\ $\mathrm{STAY}$) is in every attacker's action set, so the worst case per seed must include it.
The row ``Pessimistic, maximin'' is the full method of~\cite{sun2024belief}, and ``PB-GT, greedy'' is PB-GT. The greedy-readout medians are $7.09$ and $7.44$. Inference times are in Table~\ref{tab:speed}.
Belief-blind matched for the maximin rows, and worst-of-9/10/15/16, are Appendix~\ref{app:worst910}.}
\label{tab:gw}
\setlength{\tabcolsep}{3.5pt}
\begin{tabular}{@{}llccccc@{}}
\toprule
 & & \multicolumn{2}{c}{Clean (deployed rule)} & Clean, greedy & Matched & Worst \\
\cmidrule(lr){3-4}
Learned $Q$ & Test rule & mean${\pm}$std & median & readout & (own 3) & of 22 \\
\midrule
Vanilla & greedy & $7.35{\pm}0.24$ & $7.38$ & --- & $-612.41{\pm}166.08$ & $-968.88{\pm}49.86$ \\
\addlinespace
Pessimistic & maximin & $-687.34{\pm}305.52$ & $-788.44$ & $-15.14{\pm}70.83$ & $\mathbf{-435.30{\pm}185.98}$ & $\mathbf{-744.33{\pm}234.97}$ \\
Pessimistic & greedy & $-15.14{\pm}70.83$ & $7.09$ & --- & $-690.23{\pm}236.80$ & $-975.80{\pm}42.18$ \\
\addlinespace
PB-GT & greedy & $7.45{\pm}0.18$ & $7.44$ & --- & $-563.93{\pm}264.05$ & $-950.40{\pm}60.05$ \\
PB-GT & maximin & $-732.38{\pm}263.97$ & $-828.49$ & $7.45{\pm}0.18$ & $-610.51{\pm}235.09$ & $-848.28{\pm}167.36$ \\
\bottomrule
\end{tabular}
\end{table*}

\begin{figure*}[t]
\centering
\includegraphics[width=0.92\textwidth]{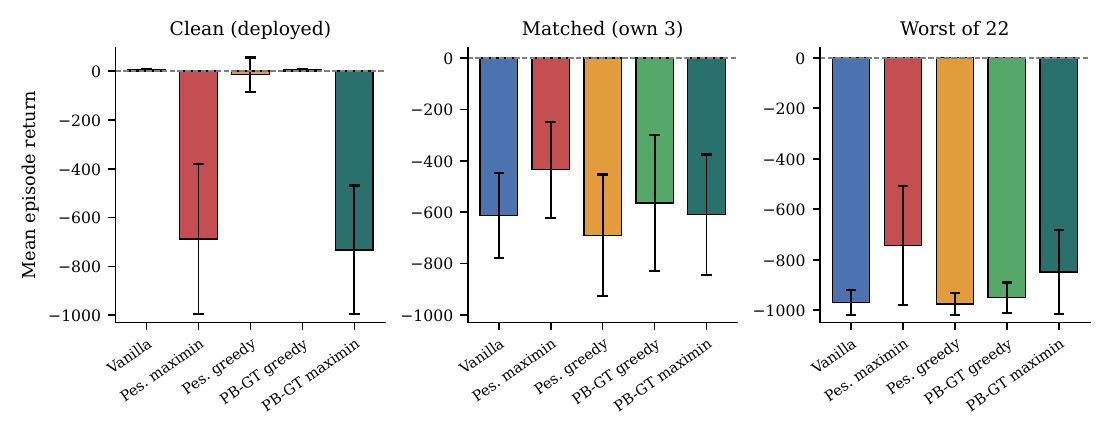}
\caption{Tabular grid, mean return $\pm$ sample std, $500$ seeds.
Clean uses the deployed rule.
Matched uses belief-aware own-three attackers for the maximin rows and belief-blind own-three for the others.
The third panel is worst-of-22.
Each axis is wide enough to include the error bars.}
\label{fig:gw-bars}
\end{figure*}

\begin{figure*}[t]
\centering
\includegraphics[width=0.95\textwidth]{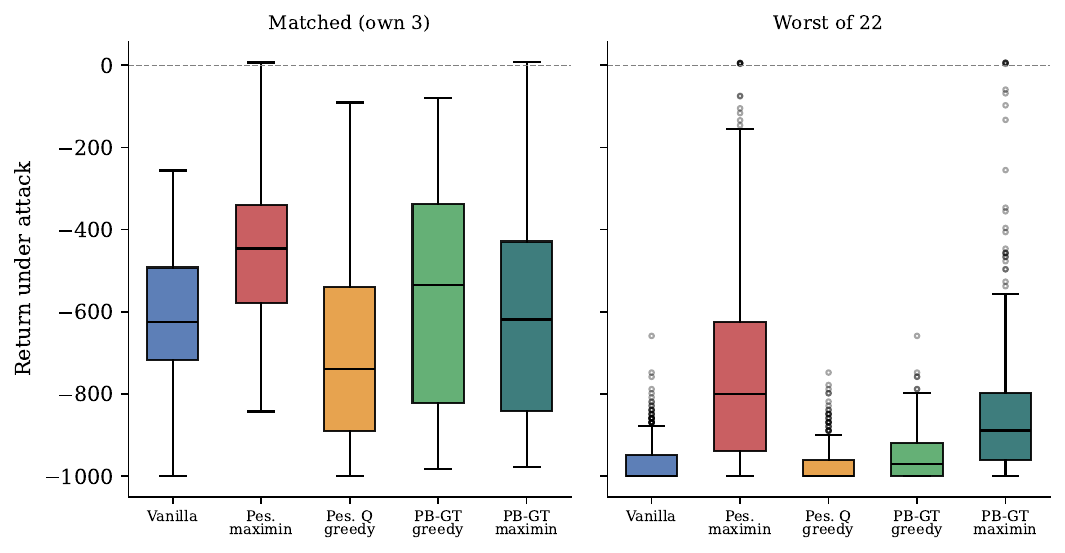}
\caption{Per-seed attack returns on the tabular GridWorld: matched (own~3) and worst-of-22.
Boxes show the median and the interquartile range, and the tails are plotted.}
\label{fig:gw-box}
\end{figure*}

Table~\ref{tab:gw} and Figures~\ref{fig:gw-bars}--\ref{fig:gw-box} are the comparison on the tabular GridWorld.
Under worst-of-22, maximin test beats greedy test on $89.6\%$ of seeds ($448$ better, $30$ tied, $22$ worse) for pessimistic $Q$ and on $76.4\%$ of seeds ($382$ / $38$ / $80$) for PB-GT $Q$.
Most of the gap between the full pessimistic agent and a greedy test rule is the test rule.

Once both agents already use maximin, the pessimistic Bellman target adds protection on most seeds against the stronger attackers, and the gain is smaller than the test-rule gain: pessimistic $Q$ with maximin test beats PB-GT $Q$ with maximin test on $79.0\%$ of seeds under belief-aware matched attackers ($395$ / $0$ / $105$; means $-435.30{\pm}185.98$ against $-610.51{\pm}235.09$) and on $65.4\%$ of seeds under worst-of-22 ($327$ / $26$ / $147$; means $-744.33{\pm}234.97$, median $-799.52$, against $-848.28{\pm}167.36$, median $-889.15$).
On the continuous GridWorld, the five-seed swap in Table~\ref{tab:cgw-swap} changes only the test rule and shows no measurable gain from the pessimistic Bellman target.

Belief-blind attackers overstated how robust maximin looks under matched evaluation.
Against their own three belief-blind attackers the maximin means were $-158.93{\pm}106.87$ (pessimistic) and $-352.12{\pm}148.23$ (PB-GT); once the attacker sees the belief those means fall to $-435.30{\pm}185.98$ and $-610.51{\pm}235.09$.
The test-rule ordering under worst-of-22 still holds.

PB-GT and vanilla, compared on each method's own three belief-blind attackers, are close.
PB-GT's matched mean is $-563.93{\pm}264.05$ and vanilla's is $-612.41{\pm}166.08$.
PB-GT is strictly better on $59.0\%$ of seeds, tied on $0.2\%$, and worse on $40.8\%$.
Under worst-of-22 the means are $-950.40{\pm}60.05$ for PB-GT and $-968.88{\pm}49.86$ for vanilla.
PB-GT is strictly better on $24.4\%$ of seeds, tied on $75.2\%$, and worse on $0.4\%$.

The belief-aware attackers also change who sets the worst case.
Under worst-of-22, counting a seed only when a single source class alone achieves the minimum, no attack alone sets the worst return on $268$ of $500$ seeds for pessimistic $Q$ with maximin test and on $159$ of $500$ for PB-GT $Q$ with maximin test; a belief-aware attacker alone sets it on $148$ and $293$ seeds respectively; one of the original fifteen alone sets it on $43$ and $6$ seeds.
Under worst-of-22 the means are $-744.33{\pm}234.97$ (median $-799.52$) for pessimistic $Q$ with maximin test, $-848.28{\pm}167.36$ (median $-889.15$) for PB-GT $Q$ with maximin test, $-975.80{\pm}42.18$ for pessimistic $Q$ with greedy test, $-950.40{\pm}60.05$ for PB-GT $Q$ with greedy test, and $-968.88{\pm}49.86$ for vanilla.
On the continuous GridWorld and on CartPole, the no-attack return is never the worst case in the suites we report; those environments are unchanged by this tabular correction.

The deployed maximin rule pays for this robustness on clean episodes.
Pessimistic clean return is $-687.34{\pm}305.52$, with median $-788.44$ and interquartile range $[-929.60,\,-547.84]$.
Maximin on PB-GT $Q$ is $-732.38{\pm}263.97$, median $-828.49$.
A $50$-seed log of the maximin-test configurations, $100$ episodes each, shows why.
Every logged episode begins with belief size $25$, the full grid, because the belief is cleared at episode boundaries.
With no attacker the observation is the true cell, the true state is inside the belief on every step, and the intersection $M\cap B_\varepsilon$ is never empty.
The mean size of that intersection is $3.64$ for pessimistic $Q$ and $3.67$ for PB-GT $Q$.
Maximin over three or four cells, on a map where the one-cell neighborhood already reaches cells next to the bomb, is cautious enough that $69.7\%$ of pessimistic clean episodes and $73.5\%$ of PB-GT clean episodes hit the $1000$-step cap.
Under attack on pessimistic $Q$ the mean intersection falls to $2.53$, still with the true state always inside and with no empty intersections.
The greedy readout of the same pessimistic $Q$, which is the clean configuration of the swapped greedy-test row, has mean $-15.14{\pm}70.83$ and median $7.09$.
The $Q$-function can still reach the goal on a typical episode when it is queried at the true cell.
The clean loss is a property of taking the worst case over a belief that has not collapsed to a singleton.

Sun and Zheng~\cite{sun2024belief} also report a clean loss when the deployed rule is used with no attack, and on Atari it is small (Pong $19.9$ against a clean $21$, Freeway $28.8$ against $34$).
They argue that no fixed policy is optimal against every attack.
A likely reason our tabular clean loss is much larger is that the budget is large relative to the map: one cell on a $5{\times}5$ grid, with neighborhoods that touch cells next to the bomb, compared with their continuous $10{\times}10$ map at $\varepsilon\in[0.1,0.5]$.
Their Theorem~1 assumes $\gamma<1$ and Lipschitz reward and transition.
Our tabular setting has $\gamma=1$ and discrete dynamics, so it does not meet those assumptions, and the tabular clean result does not contradict the theorem.
Their Table~2 ablation, maximin without a belief at $-71$ against BP-DQN at $176$ on GridWorld with $\varepsilon=0.5$, lines up with the memoryless-ball result in \S\ref{sec:cgw}: shrinking the set with history is doing work that the raw ball does not do.
A practical next step on our grid is hybrid inference that calls maximin only when the belief is still large or the margin between the top greedy actions is small, and uses the greedy readout otherwise.

\subsection{Continuous grid}
\label{sec:cgw}

\begin{table*}[t]
\centering
\small
\caption{Continuous lift of the $5{\times}5$ grid, $\varepsilon{=}0.25$, $80$k vanilla steps plus $25$k fine-tuning, $50$ episodes, $10$ seeds.
Entries are the mean $\pm$ sample std of the per-seed means.
Clean, maximin is the deployed rule for the two pessimistic rows.
Clean, greedy is the deployed rule for vanilla and PB-GT, and it is the diagnostic readout for the pessimistic rows.
PB-GT was not evaluated with a maximin test on all ten seeds.}
\label{tab:cgw}
\setlength{\tabcolsep}{4pt}
\begin{tabular}{@{}lcccc@{}}
\toprule
Condition & Vanilla & Pes.\ (box) & PB-GT & Pes.\ (ball) \\
\midrule
Clean, greedy & $4.08{\pm}1.17$ & $4.51{\pm}0.47$ & $4.60{\pm}0.46$ & $0.58{\pm}2.44$ \\
Clean, maximin & --- & $3.67{\pm}2.11$ & --- & $-0.55{\pm}2.43$ \\
Shared-random & $4.34{\pm}0.59$ & $4.37{\pm}0.37$ & $4.24{\pm}0.69$ & $0.26{\pm}1.92$ \\
PGD-10 & $4.06{\pm}0.68$ & $4.02{\pm}0.46$ & $4.06{\pm}0.58$ & $-2.86{\pm}2.64$ \\
Matched search & $-20.75{\pm}9.09$ & $1.98{\pm}2.94$ & $-16.12{\pm}9.43$ & $-26.02{\pm}14.26$ \\
Matched SB3 & $-18.12{\pm}9.88$ & $1.34{\pm}2.59$ & $-14.06{\pm}8.48$ & $-28.22{\pm}15.68$ \\
\bottomrule
\end{tabular}
\end{table*}

\begin{figure*}[t]
\centering
\includegraphics[width=\textwidth]{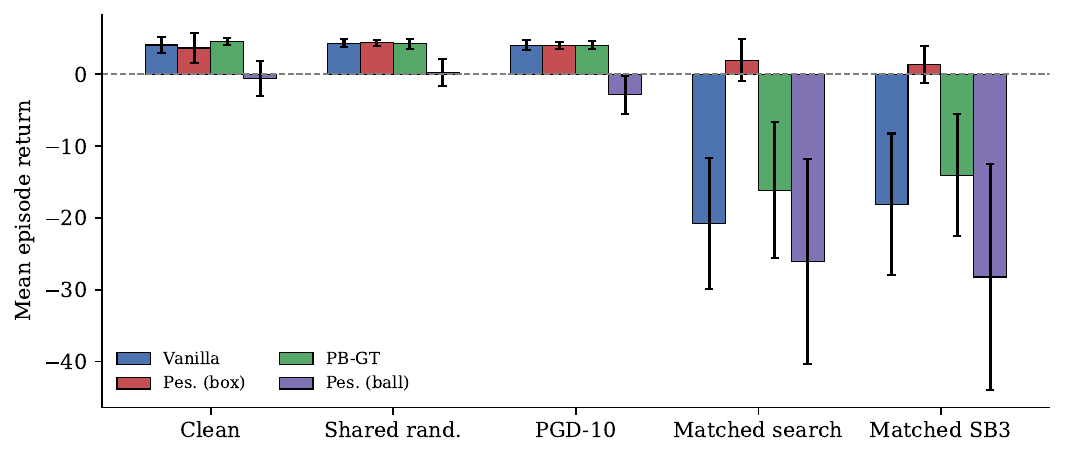}
\caption{Continuous GridWorld, ten seeds, $\varepsilon{=}0.25$.
Clean bars use the deployed rule: maximin for the two pessimistic methods, greedy for vanilla and PB-GT.
Error bars are the sample std of the per-seed means.}
\label{fig:cgw}
\end{figure*}

\begin{table}[t]
\centering
\small
\caption{Target/test swap on the continuous GridWorld, seeds $0$--$4$ only.
These five seeds are a subset of Table~\ref{tab:cgw}.
Entries are mean $\pm$ sample std of the per-seed means.
Diagonal entries are seeds $0$--$4$ of the ten-seed comparison.
Off-diagonal PGD-10 entries are the same seeds under the swapped test rule.
The off-diagonal entries change only the test rule (Table~\ref{tab:swap}).}
\label{tab:cgw-swap}
\setlength{\tabcolsep}{3pt}
\begin{tabular}{@{}lcc@{}}
\toprule
 & Pes.\ $Q$ & PB-GT $Q$ \\
\midrule
\multicolumn{3}{@{}l}{\emph{Clean (deployed rule)}} \\
Maximin test & $2.85{\pm}2.86$ & $4.41{\pm}0.37$ \\
Greedy test & $4.42{\pm}0.29$ & $4.62{\pm}0.47$ \\
\multicolumn{3}{@{}l}{\emph{Matched search}} \\
Maximin test & $0.82{\pm}3.14$ & $1.86{\pm}2.46$ \\
Greedy test & $-15.44{\pm}4.07$ & $-17.01{\pm}7.84$ \\
\multicolumn{3}{@{}l}{\emph{Matched SB3}} \\
Maximin test & $-0.07{\pm}2.78$ & $2.26{\pm}3.69$ \\
Greedy test & $-15.05{\pm}6.77$ & $-17.63{\pm}9.98$ \\
\multicolumn{3}{@{}l}{\emph{PGD-10}} \\
Maximin test & $3.78{\pm}0.36$ & $4.42{\pm}0.38$ \\
Greedy test & $0.96{\pm}2.71$ & $3.91{\pm}0.72$ \\
\bottomrule
\end{tabular}
\end{table}

Table~\ref{tab:cgw} and Figure~\ref{fig:cgw} are the ten-seed comparison on the continuous GridWorld.
With the deployed maximin rule, pessimistic(box) clean return is $3.67{\pm}2.11$, milder than the tabular clean loss, and the greedy readout of the same weights is $4.51{\pm}0.47$, next to vanilla's greedy clean return of $4.08{\pm}1.17$.
The memoryless ball does not keep a usable clean policy (maximin $-0.55{\pm}2.43$, greedy readout $0.58{\pm}2.44$).
Pessimistic(box) is strictly above vanilla on all $10$ seeds under matched search and on all $10$ seeds under matched SB3.
On the other suites the strict-win counts are $3/10$ on deployed clean, $4/10$ on shared-random, and $2/10$ on PGD-10.
PB-GT's matched means, $-16.12{\pm}9.43$ and $-14.06{\pm}8.48$, stay in the same range as vanilla's $-20.75{\pm}9.09$ and $-18.12{\pm}9.88$, far below the box.

PGD-10 barely moves vanilla, the box, or PB-GT.
It is a standard MinBest attack, and on this environment it flips the greedy action on $28\%$ of visited states ($47\%$ of states where some lattice point inside the ball would flip the action), which is why it is weak here.

Table~\ref{tab:cgw-swap} is that swap on seeds $0$--$4$ only.
It takes each trained $Q$-function and changes only the test rule.
On both matched suites, maximin test beats greedy test on all five seeds for the pessimistic $Q$-network and on all five seeds for the PB-GT network.
The clean column of that table is the other direction on some seeds: the deployed maximin reading of pessimistic $Q$ averages $2.85{\pm}2.86$, while the greedy reading averages $4.42{\pm}0.29$.
PB-GT $Q$ with the maximin test is higher on average than pessimistic $Q$ with the maximin test on clean ($4.41{\pm}0.37$ against $2.85{\pm}2.86$), on matched search ($1.86{\pm}2.46$ against $0.82{\pm}3.14$), and on matched SB3 ($2.26{\pm}3.69$ against $-0.07{\pm}2.78$); per seed, $2$ of $5$ on matched search and $5$ of $5$ on clean and matched SB3.
On the continuous GridWorld, the pessimistic Bellman target adds no measurable gain in these five seeds over the test rule.
Five seeds are enough to see the matched-attack ordering.
They are not a substitute for the tabular study of heavy tails.

\subsection{CartPole: trained sound box}
\label{sec:cp}

\begin{table*}[t]
\centering
\small
\caption{Trained sound-box CartPole, $\varepsilon{=}0.05$, $50$ episodes, five seeds.
Mean $\pm$ sample std of the per-seed means.
Clean for pessimistic is the deployed maximin rule.
The shaped attacker uses reward $|\theta|+0.1|x|$.}
\label{tab:cp}
\setlength{\tabcolsep}{6pt}
\begin{tabular}{@{}lcccc@{}}
\toprule
Method & Clean & Shared-random & PGD* & \shortstack{Shaped\\$|\theta|{+}0.1|x|$} \\
\midrule
Vanilla & $500.0{\pm}0.0$ & $290.9{\pm}181.3$ & $300.1{\pm}163.8$ & $300.2{\pm}181.6$ \\
Pessimistic & $500.0{\pm}0.0$ & $483.5{\pm}37.0$ & $459.0{\pm}91.6$ & $420.0{\pm}178.9$ \\
PB-GT & $418.3{\pm}163.1$ & $131.3{\pm}131.4$ & $133.7{\pm}82.7$ & $258.6{\pm}184.9$ \\
\bottomrule
\end{tabular}
\end{table*}

\begin{table*}[t]
\centering
\small
\caption{Per-seed returns for Table~\ref{tab:cp}, rounded to the nearest integer, in the order vanilla / pessimistic / PB-GT.
A superscript $v$ marks a vacuous shaped configuration.
The last column is the selected pessimistic checkpoint.}
\label{tab:cp-seeds}
\setlength{\tabcolsep}{4pt}
\begin{tabular}{@{}cccccc@{}}
\toprule
Seed & Clean & Shared-random & PGD* & Shaped & Pes.\ step \\
\midrule
$0$ & $500/500/464$ & $215/500/74$ & $287/500/136$ & $234/500/56$ & $5000$ \\
$1$ & $500/500/500$ & $152/500/15$ & $259/500/50$ & $26/500^{v}/376$ & $50000$ \\
$2$ & $500/500/500$ & $500/500/287$ & $500/500/70$ & $500^{v}/500^{v}/500$ & $50000$ \\
$3$ & $500/500/500$ & $116/500/259$ & $62/500/153$ & $406/500/257$ & $50000$ \\
$4$ & $500/500/128$ & $471/417/23$ & $393/295/259$ & $335/100/105$ & $50000$ \\
\bottomrule
\end{tabular}
\end{table*}

Table~\ref{tab:cp} is the trained CartPole comparison, and Table~\ref{tab:cp-seeds} is the same returns seed by seed.
Vanilla's clean return is the greedy policy, and it is $500.0$ on every seed.
Pessimistic's deployed maximin clean return is $500.0$ on every seed.
PB-GT's clean mean is $418.3{\pm}163.1$.
Under shared-random the means are $290.9{\pm}181.3$, $483.5{\pm}37.0$, and $131.3{\pm}131.4$.
Under PGD* they are $300.1{\pm}163.8$, $459.0{\pm}91.6$, and $133.7{\pm}82.7$.
Seed~$2$ ties vanilla at $500$ on shared-random and on PGD*.
Seed~$4$ is the loss, on shared-random ($417$ against vanilla $471$), on PGD* ($295$ against $393$), and on the shaped attacker.

Against the shaped attacker the per-seed comparison of pessimistic with vanilla is $2$ non-vacuous wins, $1$ vacuous win, $1$ tie, and $1$ loss.
The non-vacuous wins are seeds $0$ and $3$.
The vacuous win is seed~$1$, where pessimistic scores $500$ and the shaped attacker never ended an episode.
The tie is seed~$2$, where both agents score $500$ and both configurations are vacuous.
The loss is seed~$4$.
The unshaped matched attacker is vacuous against pessimistic on $4$ of $5$ seeds.
Seed~$4$, the one non-vacuous configuration, returns $500$, the same value vanilla returns on that seed.
We do not treat that column as evidence of robustness, and it is omitted from Table~\ref{tab:cp}.

The selected pessimistic checkpoint is step $5000$ on seed~$0$ and step $50000$ on seeds $1$--$4$.
On seed~$0$ the $10$k evaluation, $15$ episodes, read $234.9$ under a greedy readout and $312.9$ under deployed maximin.
Training stopped and the weights rolled back to step $5000$, where both reads were $500$.
The ordering is fragile above $\varepsilon{=}0.05$.
On the two-seed pole-sensor sweep in Appendix~\ref{app:cp-extra}, the sound box holds $500$ at $\varepsilon\in\{0.025,0.05\}$, and at $\varepsilon{=}0.10$ the worst-of-$9$ sound mean is $139.9$ (greedy $10.6$, uniform $74.0$).
At $\varepsilon\in\{0.15,0.20\}$ the worst-of-$9$ sound mean is $9.1$.

The same sound box on frozen vanilla weights, with no fine-tuning, holds $500$ under the deployed maximin rule on clean episodes, under shared-random, and under PGD* (Table~\ref{tab:cp-sound}).
Those two attacks drop the greedy readout of the same weights to $290.9{\pm}181.3$ and $300.1{\pm}163.8$.
The frozen result supports the test-rule reading of the trained table: the gain is available from the decision rule on weights that were not fine-tuned under attack.

\subsection{Inference cost}
\label{sec:speed}

\begin{figure}[t]
\centering
\includegraphics[width=\columnwidth]{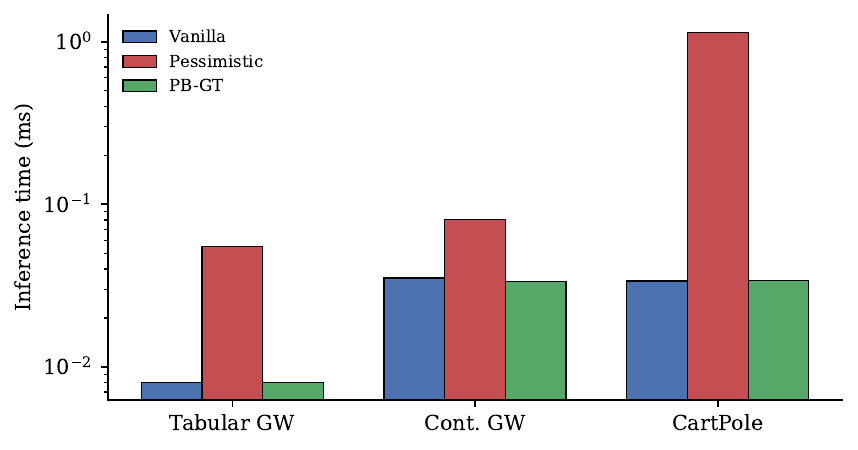}
\caption{Single-process inference time of the deployed rules, log scale.
Tabular times are the attack path on $20$ seeds.
Continuous times are single-process timings of the deployed box on the ten seeds.
CartPole times are the trained sound box.}
\label{fig:speed}
\end{figure}

\begin{table}[t]
\centering
\small
\caption{Single-process inference, milliseconds per action, mean $\pm$ sample std across seeds.
Tabular: attack path of the deployed rules, $n{=}20$.
Continuous: single-process timings of the deployed box, $n{=}10$.
CartPole: mean of the five per-seed means.}
\label{tab:speed}
\setlength{\tabcolsep}{4pt}
\begin{tabular}{@{}lccc@{}}
\toprule
Environment & Vanilla & Pessimistic & PB-GT \\
\midrule
Tabular & $0.0080{\pm}0.0005$ & $0.0548{\pm}0.0024$ & $0.0080{\pm}0.0004$ \\
Continuous & $0.0352{\pm}0.0069$ & $0.0810{\pm}0.0046$ & $0.0336{\pm}0.0008$ \\
CartPole & $0.0337{\pm}0.0006$ & $1.14{\pm}0.05$ & $0.0339{\pm}0.0018$ \\
\bottomrule
\end{tabular}
\end{table}

Table~\ref{tab:speed} and Figure~\ref{fig:speed} are the timing comparison.
On the tabular GridWorld the deployed maximin path costs $0.0548$\,ms against $0.0080$\,ms for vanilla, a factor of $6.8$.
PB-GT matches the vanilla time.
On the continuous GridWorld the box costs $0.0810$\,ms against $0.0352$\,ms for vanilla, a factor of $2.3$, and PB-GT is $0.0336$\,ms.
On the trained CartPole box the times are $0.0337$, $1.14$, and $0.0339$\,ms, about $34$ times for the maximin over the $16$-point lattice.
These factors are the cost of the searches in \S\ref{sec:pes}.
A finer lattice would cost more.

\section{Related work}
\label{sec:related}

Huang et al.~\cite{huang2017adversarial} transferred FGSM-style attacks from supervised learning to Atari policies.
Zhang et al.~\cite{zhang2020robust} formulated the state-adversarial MDP and the SA-DQN regularizer, and a follow-up trained the optimal adversary as an RL agent (ATLA)~\cite{zhang2021robust}.
Liang et al.~\cite{liang2022wocar} estimate worst-case values without an explicit attacker during training (WocaR-RL).
We train frozen DQN attackers for evaluation, following the attacker-as-MDP view, and we do not co-train them with the defender.
Ilahi et al.~\cite{ilahi2022challenges} survey the broader attack literature.
Sun, Liu, Ding, and Zheng~\cite{sun2025shift} show that $\ell_p$-bounded attacks often fail to change the semantics of an image, and they propose a diffusion attack that does.
Our threat model stops at the $\varepsilon$-ball.
Those semantic attacks are outside it, and nothing in the tables speaks to them.

The direct predecessor is Sun and Zheng~\cite{sun2024belief}: the Stackelberg formulation, the maximin rule, the tabular belief, and BP-DQN / DP-DQN.
Our tabular agent follows their Algorithms~1--2.
The deep experiments use the sound box of Appendix~\ref{app:box}.
The belief network in BP-DQN is a particle-filter RNN~\cite{ma2020pfrnn}, which we do not train.
CAR-DQN~\cite{li2024car} minimizes a surrogate of the Bellman infinity-error and keeps greedy inference.
We have not run it, or SA-DQN, WocaR-RL, ATLA, or Atari.

\section{Limitations}
\label{sec:limit}

\paragraph{Clean cost on the tabular GridWorld.}
The deployed maximin rule does poorly with no attacker on our $5{\times}5$ grid (\S\ref{sec:gw}).
Our explanation, that a one-cell budget is large relative to this map, is untested; a larger grid would check it.
CartPole shows no clean cost, and the continuous GridWorld a mild one.

\paragraph{Attackers.}
The tabular attackers were trained against three defender seeds and reused across all $500$, so, as defined in \S\ref{sec:gw-setup}, matched means matched to the rule rather than to each agent.
On continuous GridWorld and CartPole the learned SB3 attackers do not observe the box; continuous matched search does score under the defender's actual rule.
On CartPole the unshaped attacker found no effective attack against the pessimistic agent on $4$ of $5$ seeds (\S\ref{sec:cp}).

\paragraph{Scope.}
This is not a reproduction of BP-DQN: the deep experiments use a sound box built from the true dynamics instead of a learned belief, and we did not run their $10{\times}10$ maze, particle-filter RNN, or Atari.
The deep results rest on fewer seeds ($10$ on the continuous GridWorld with a five-seed swap, five on CartPole), a single continuous budget, and a CartPole ordering that weakens above $\varepsilon{=}0.05$.
We did not compare against CAR-DQN, SA-DQN, WocaR-RL, or ATLA.

\section{Conclusion}
\label{sec:conc}

The robustness of belief-enriched pessimistic Q-learning, in the settings we ran, is mostly in the test-time maximin over the belief.
On the tabular GridWorld, switching the test rule from greedy to maximin wins on $89.6\%$ of seeds for pessimistic $Q$ and on $76.4\%$ for PB-GT $Q$ under worst-of-22, and the pessimistic Bellman target adds protection on most seeds against the stronger attackers ($79.0\%$ belief-aware matched, $65.4\%$ worst-of-22), a smaller gain than the test-rule switch.
PB-GT performs about like vanilla under attack, so adversarial data alone does not buy the robustness.
On the continuous GridWorld, the five-seed swap, which changes only the test rule, shows no measurable further gain from the pessimistic Bellman target.
The frozen CartPole box holds $500$ under shared random noise and under PGD*, a random search inside the budget, on weights that were never fine-tuned with the box.
The lookup costs about $6.8\times$ on the tabular GridWorld, about $2.3\times$ on the continuous GridWorld, and about $34\times$ on CartPole.
Each factor is for the search used in that experiment (\S\ref{sec:pes}).
On the tabular GridWorld the same rule is weak with no attacker, because the belief stays a few cells wide; the useful follow-up is to spend the maximin only when that belief, or the greedy margin, says the observation is ambiguous.

\appendix

\section{The sound-box belief}
\label{app:box}

This appendix describes the belief used on continuous GridWorld and on CartPole.
It is an axis-aligned box built from the known dynamics.
It is not a particle set and it does not use a recurrent belief network.

On the continuous GridWorld the state is $(x,y)$ and each action translates the agent by $0.5$ unless a wall blocks the move.
Given a box $M_t$ and an observation $\tilde s_t$, we first propagate $M_t$ through the translation.
If a wall makes the image of the box set-valued, we replace the box by the axis-aligned hull of the box and its image.
We then intersect the propagated box with $B_\varepsilon(\tilde s_{t+1})$, and we clip to the domain $[0,5]^2$.
The inner min is a hard minimum of $Q$ over a $3{\times}3$ lattice on that intersected box.
A memoryless ball, used as an ablation, skips the propagation and places the same lattice on $B_\varepsilon(\tilde s)$ itself.
Both sets contain the true state when the observation respects the $\ell_\infty$ budget.
Only the box can shrink when later observations cut it.

On CartPole the observation is $(x,\dot x,\theta,\dot\theta)$, and the dynamics are the standard cart-pole equations, applied as interval arithmetic on the box.
The update filters first and then propagates.
After an observation arrives, coordinates that the $\varepsilon$-ball rules out are cut away.
The surviving box is then pushed one step through the interval dynamics under the action that was executed, and the next observation cuts it again.
The inner min is a hard minimum over a $16$-point lattice, two samples along each coordinate of the box.
When the intersection is computed correctly the true state stays inside the box.
That is the soundness property of \S\ref{sec:belief}.

\section{Attacker protocol and hyperparameters}
\label{app:hyper}

\paragraph{Tabular attackers.}
The evaluation set has twenty-one Stable-Baselines3 DQN attackers.
Fifteen are belief-blind: three for each of five rules (vanilla, pessimistic $Q$ with maximin test, PB-GT $Q$ with greedy test, pessimistic $Q$ with greedy test, and PB-GT $Q$ with maximin test).
Six are belief-aware: three for pessimistic $Q$ with maximin test and three for PB-GT $Q$ with maximin test.
For each triple the attackers are trained against defender seeds $99999$, $99998$, and $99997$, each with its own attacker RNG, for $2{\times}10^5$ steps.
The attacker's policy is an MLP with hidden sizes $[256,128]$, learning rate $10^{-3}$, and $\gamma{=}1$.
Its action is a grid direction, clipped so that $\tilde s\in B_\varepsilon(s)$.
A belief-aware attacker observes the true cell and a $25$-dimensional binary belief mask; a belief-blind attacker observes only the true cell.
Matched return in the main table averages a rule's own three attackers (belief-aware for maximin rows, belief-blind otherwise).
Worst-of-22 takes the per-seed minimum over all twenty-one attackers and the clean return.
Vanilla's matched column uses its own three.
Vanilla's worst-of-15 is the per-seed minimum of the original nine returns and the returns against the six swapped-rule attackers.

\paragraph{Tabular GridWorld.}
The grid is $5{\times}5$, with reward $-1$ per step, $+10$ on gold, and $-10$ on the bomb.
We use $\gamma{=}1$.
The learning rate $\alpha$ is annealed from $0.1$ to $0.005$ for vanilla and from $0.1$ to $0.01$ for the fine-tunes.
During fine-tuning, $\eta$-greedy exploration is annealed from $0.2$ to $0.01$.
Vanilla trains for $2000$ episodes.
Pessimistic Q-learning and PB-GT each fine-tune for $500$ episodes from that table.
Evaluation uses $100$ episodes and a cap of $1000$ steps, repeated for seeds $0$--$499$.
Both the pessimistic policy and its Bellman target use a hard min over the belief.
The PB-GT Bellman target is greedy Q-learning on the true next state.

\paragraph{Continuous GridWorld.}
The domain is $[0,5]^2$ with step size $0.5$.
The network is a $2{\to}128{\to}128{\to}4$ MLP trained with Double DQN, Huber loss, and Adam.
The learning rate is $10^{-3}$ for vanilla and $10^{-4}$ for fine-tuning.
We use $\gamma{=}0.99$, a replay buffer of $10^5$, batch size $64$, learning starts at $10^3$ steps, and the target network is copied every $500$ steps.
Vanilla trains for $80$k steps.
Pessimistic and PB-GT fine-tune for $25$k steps from the same vanilla weights.
The budget in the production table is $\varepsilon{=}0.25$ under $L_\infty$.
The inner min is a hard min over a $3{\times}3$ lattice.
Evaluation uses $50$ episodes, $10$ seeds for the main table, and a cap of $100$ steps.
The swap table uses seeds $0$--$4$.
PGD-10 is a $10$-step iterative MinBest attack.
Matched search uses $32$ candidates.
Matched SB3 trains for $40$k steps with hidden sizes $[128,128]$ and a $5{\times}5$ perturbation grid.

\paragraph{CartPole.}
The vanilla network has hidden sizes $[128,128]$ and ReLU activations.
We use Adam at $10^{-3}$, $\gamma{=}0.99$, a buffer of $10^5$, batch size $128$, learning starts at $10^3$ steps, a gradient step every $4$ environment steps, and a target copy every $10$ episodes.
Exploration $\eta$ is annealed from $1.0$ to $0.01$ over $50$k steps.
The loss is Huber with gradient clipping at $10$.
Training runs for $5{\times}10^5$ steps.
The trained sound-box fine-tunes freeze the backbone and train the last linear layer at $10^{-5}$.
The inner min uses $16$ lattice points on the interval box.
Pessimistic fine-tuning is capped at $5{\times}10^4$ steps and is scored every $5{\times}10^3$ steps, including step~$0$, by clean return under deployed maximin.
Training stops early if that return has reached $475$ and a later evaluation falls below $400$.
PB-GT runs for $4{\times}10^4$ steps with greedy Double DQN targets.
Evaluation uses $\varepsilon{=}0.05$, $50$ episodes, and five seeds.
PGD* draws $32$ candidates.
Unshaped and shaped SB3 attackers train for $40$k steps.
The shaped reward is $|\theta|+0.1|x|$.
Checkpoint selection is in \S\ref{sec:cp}.

\section{Additional tabular attack summaries}
\label{app:worst910}

Table~\ref{tab:gw-w910} reports belief-blind matched means for the maximin rows, together with worst-of-9, worst-of-10, worst-of-15, and worst-of-16, on the same $500$ seeds as Table~\ref{tab:gw}.
Worst-of-9 is the original nine attackers; worst-of-10 adds no attack.
Worst-of-15 is all fifteen belief-blind attackers; worst-of-16 adds no attack.
For vanilla, pessimistic greedy, and PB-GT greedy, worst-of-9 agrees with worst-of-10 and worst-of-15 agrees with worst-of-16, because clean return is never the minimum there.
Under worst-of-9, PB-GT greedy beats vanilla on $27.8\%$ of seeds, ties on $70.0\%$, and loses on $2.2\%$; under worst-of-10 the same comparison is unchanged for those rows.

\begin{table*}[t]
\centering
\small
\caption{Tabular grid, $500$ seeds: belief-blind matched (own~3) for the maximin rows, plus worst-of-9/10/15/16.
Mean $\pm$ sample std (ddof${=}1$).
A dash means the column is not the headline matched metric for that row (greedy and vanilla matched stay in Table~\ref{tab:gw}).}
\label{tab:gw-w910}
\setlength{\tabcolsep}{3.5pt}
\begin{tabular}{@{}lccccc@{}}
\toprule
Rule & Belief-blind matched & Worst-of-9 & Worst-of-10 & Worst-of-15 & Worst-of-16 \\
\midrule
Vanilla & --- & $-950.67{\pm}65.09$ & $-950.67{\pm}65.09$ & $-968.86{\pm}49.87$ & $-968.86{\pm}49.87$ \\
Pes.\ $Q$, maximin & $-158.93{\pm}106.87$ & $-319.92{\pm}126.29$ & $-702.61{\pm}276.39$ & $-495.79{\pm}173.47$ & $-719.16{\pm}251.38$ \\
Pes.\ $Q$, greedy & --- & $-968.64{\pm}45.79$ & $-968.64{\pm}45.79$ & $-975.80{\pm}42.18$ & $-975.80{\pm}42.18$ \\
PB-GT $Q$, greedy & --- & $-924.36{\pm}82.48$ & $-924.36{\pm}82.48$ & $-950.30{\pm}60.27$ & $-950.30{\pm}60.27$ \\
PB-GT $Q$, maximin & $-352.12{\pm}148.23$ & $-377.43{\pm}132.71$ & $-753.24{\pm}229.41$ & $-542.93{\pm}190.87$ & $-763.42{\pm}216.26$ \\
\bottomrule
\end{tabular}
\end{table*}

\section{Code and data}
\label{app:code}

Code and logs: \url{https://github.com/adithya-venkatesh/RL}.

{\raggedright\footnotesize
\begin{itemize}[leftmargin=1.1em]
\item Tabular $2{\times}2$, $500$ seeds, fifteen belief-blind attackers:
\texttt{gridworld\_target\_test\_swap\_tabular\_matched15\_v3.csv}.
\item Worst-of-16 (belief-blind attacks plus no attack), with summary and win-rate sidecars:
\texttt{gridworld\_worst\_of\_16\_v3.csv}.
\item Belief-aware attackers and worst-of-22, with summary and win-rate sidecars:
\texttt{gridworld\_belief\_aware\_attack\_v3.csv}
(\texttt{gridworld\_belief\_aware\_attack\_v3.py},
\texttt{gridworld\_belief\_aware\_common\_v3.py}).
\item Vanilla clean return, matched return on its own three attackers, and the original nine attack returns:
\texttt{gridworld\_experiments\_500runs\_final\_v3.csv}.
Vanilla against the six swapped-rule attackers, and the worst-of-15 minimum:
\texttt{gridworld\_vanilla\_vs\_swap\_attackers\_v3.csv}.
\item Belief size and clean truncation, $50$ seeds, $100$ episodes:
\texttt{gridworld\_clean\_maximin\_belief\_v3.csv}.
\item Tabular attack-path timing, $20$ seeds:
\texttt{gridworld\_inference\_20seeds\_final\_v3.csv}.
\item Continuous production comparison, ten seeds, $\varepsilon{=}0.25$:
\texttt{gw\_prod\_eps025\_10seeds\_v3.csv}.
\item Continuous swap, seeds $0$--$4$:
\texttt{gridworld\_target\_test\_swap\_cgw\_v3.csv}.
PGD-10 under the swapped test rule:
\texttt{gridworld\_target\_test\_swap\_cgw\_pgd\_offdiag\_v3.csv}.
\item Single-process timings.
Continuous rows are the deployed box.
Tabular rows are greedy clean readouts and are not the maximin cost in Table~\ref{tab:speed}:
\texttt{inference\_retime\_singleproc\_v3.csv}.
\item Trained CartPole sound box, five seeds:
\texttt{cartpole\_trained\_box\_5seeds\_v3.csv},
checkpoint curve \texttt{cartpole\_trained\_box\_5seeds\_curves\_v3.csv}.
\item Frozen CartPole probes:
\texttt{cartpole\_sound\_5seeds.csv},
\texttt{cartpole\_sound\_bias5seeds.csv},
\texttt{cartpole\_sound\_epssweep\_v2.csv},
\texttt{cartpole\_sound\_possensor\_5seeds.csv},
and the one-seed pole-sensor probe \texttt{cartpole\_sound\_epssweep.csv}.
\item Continuous budget sweeps:
\texttt{gw\_epssweep\_eps025.csv},
\texttt{gw\_epssweep\_eps030.csv},
\texttt{gw\_epssweep\_eps035.csv},
\texttt{gw\_epssweep\_eps040.csv},
\texttt{gw\_epssweep\_eps050.csv},
\texttt{gw\_epssweep\_ft25k\_eps030.csv},
\texttt{gw\_epssweep\_ft25k\_eps035.csv},
\texttt{gw\_epssweep\_ft25k\_eps040.csv}.
\item GRU-particle port that did not hold: \texttt{cartpole\_experiments\_restore\_5runs.csv}.
\item Figures: \texttt{arxiv\_v1/build\_figures.py}.
\end{itemize}}
Reference code for BP-DQN and DP-DQN: \url{https://github.com/SliencerX/Belief-enriched-robust-Q-learning}.
CAR-DQN: \url{https://github.com/leoranlmia/CAR-DQN}.

\section{PB-GT DQN pseudocode}
\label{app:alg}

The tabular loops are Algorithms~\ref{alg:pes-tab} and~\ref{alg:pbgt-tab}.
The deep loop uses the sound box of Appendix~\ref{app:box}.

\paragraph{Training.}
Start from a pretrained vanilla network $Q_v$ and a copy $Q'$ used as the target network.
On CartPole, freeze every layer except the last linear layer.
For each step, draw an attacked observation $\tilde s_t\in B_\varepsilon(s_t)$ with the training attack (the white-box objective in the tabular case, a random-search perturbation in these deep runs).
Form the box $M_t$ by the update in Appendix~\ref{app:box}.
With probability $1-\eta$ take $a_t=\arg\max_a\min_{m\in M_t}Q_g(m,a)$, and otherwise take a random action.
Execute $a_t$, observe the reward and the true next state $s_{t+1}$, and store $(s_t,a_t,R_t,s_{t+1})$.
The box may be stored with the transition for bookkeeping.
The target does not read it.
The target is the Double DQN greedy target on the true next state,
\[
y=
\begin{cases}
R & \text{if $s'$ is terminal,}\\
R+\gamma\,Q'\bigl(s',\,\arg\max_{a'}Q_g(s',a')\bigr) & \text{otherwise.}
\end{cases}
\]
Take a gradient step on the Huber loss and copy $Q_g$ into $Q'$ on the usual period.
Do not use a maximin target.

\paragraph{Testing.}
Observe $\tilde s_t$ and take $a_t=\arg\max_a Q_g(\tilde s_t,a)$.
There is no box and no maximin at test time.

\section{Budgets that were swept}
\label{app:eps}

Table~\ref{tab:cgw} is the $80$k vanilla plus $25$k fine-tune schedule at $\varepsilon{=}0.25$, ten seeds, and $50$ episodes.
That final schedule was run only at $\varepsilon{=}0.25$.
$\varepsilon{=}0.25$ was chosen from the two shorter sweeps below, where the pessimistic greedy clean return is least damaged.
In Table~\ref{tab:eps-100k} that column is highest at $0.25$.
The deployed maximin clean return in the same table is negative at every budget and closest to zero at $0.30$.
Table~\ref{tab:eps-40k} does not include $0.25$.
Neither sweep is the production comparison.

\begin{table}[t]
\centering
\scriptsize
\setlength{\tabcolsep}{3pt}
\caption{$100$k vanilla steps, $40$k fine-tuning, $30$ episodes, seeds $0$--$2$.
Clean return, mean $\pm$ sample std across seeds.
Pessimistic maximin is the deployed box rule.}
\label{tab:eps-100k}
\begin{tabular}{@{}lcccc@{}}
\toprule
$\varepsilon$ & Vanilla & \shortstack{Pes.\\maximin} & \shortstack{Pes.\\greedy} & PB-GT \\
\midrule
$0.25$ & $4.46{\pm}0.24$ & $-3.14{\pm}3.45$ & $4.03{\pm}0.44$ & $4.46{\pm}0.24$ \\
$0.30$ & $4.46{\pm}0.24$ & $-0.07{\pm}0.25$ & $2.60{\pm}1.42$ & $4.43{\pm}0.26$ \\
$0.35$ & $4.46{\pm}0.24$ & $-4.57{\pm}4.00$ & $-0.74{\pm}3.28$ & $4.46{\pm}0.24$ \\
$0.40$ & $4.46{\pm}0.24$ & $-21.23{\pm}20.84$ & $-6.81{\pm}9.61$ & $4.43{\pm}0.23$ \\
$0.50$ & $4.46{\pm}0.24$ & $-10.92{\pm}8.92$ & $-3.23{\pm}3.67$ & $2.24{\pm}3.94$ \\
\bottomrule
\end{tabular}
\end{table}

\begin{table}[t]
\centering
\scriptsize
\setlength{\tabcolsep}{3pt}
\caption{$40$k vanilla steps, $25$k fine-tuning, $20$ episodes, seeds $0$--$1$.
Clean return, mean $\pm$ sample std across the two seeds.}
\label{tab:eps-40k}
\begin{tabular}{@{}lcccc@{}}
\toprule
$\varepsilon$ & Vanilla & \shortstack{Pes.\\maximin} & \shortstack{Pes.\\greedy} & PB-GT \\
\midrule
$0.30$ & $4.58{\pm}0.11$ & $-1.50{\pm}6.36$ & $3.82{\pm}0.74$ & $4.58{\pm}0.11$ \\
$0.35$ & $4.58{\pm}0.11$ & $1.92{\pm}1.73$ & $1.03{\pm}0.81$ & $4.58{\pm}0.11$ \\
$0.40$ & $4.58{\pm}0.11$ & $-8.12{\pm}5.98$ & $2.33{\pm}1.10$ & $4.58{\pm}0.11$ \\
\bottomrule
\end{tabular}
\end{table}

The CartPole pole-sensor sweeps recorded attack return, not clean return.
Seed~$0$ alone, $3$ episodes per configuration, covered $\varepsilon\in\{0.05,0.10,0.20\}$.
Seeds $0$ and $1$, $20$ episodes per configuration, covered $\varepsilon\in\{0.025,0.05,0.10,0.15,0.20\}$.
Table~\ref{tab:cp-eps} is the mean of those two seeds under the worst of nine constant biases.
The same two seeds under an exhaustive search of the nine directions have sound returns $500.0$, $500.0$, $190.6$, $86.6$, and $29.9$ at those five budgets, with greedy $141.2$, $488.0$, $500.0$, $371.2$, $14.8$ and uniform $470.1$, $462.0$, $133.2$, $13.8$, $14.9$.
On the one-seed run the worst-of-nine sound returns are $500.0$, $134.0$, and $8.7$, with greedy $179.0$, $10.7$, $34.7$ and uniform $500.0$, $61.0$, $8.7$.
Its exhaustive sound returns are $500.0$, $180.3$, and $21.7$, with greedy $500.0$, $500.0$, $10.0$ and uniform $500.0$, $235.0$, $15.0$.

\section{CartPole probes outside the trained comparison}
\label{app:cp-extra}

\begin{figure}[t]
\centering
\includegraphics[width=\columnwidth]{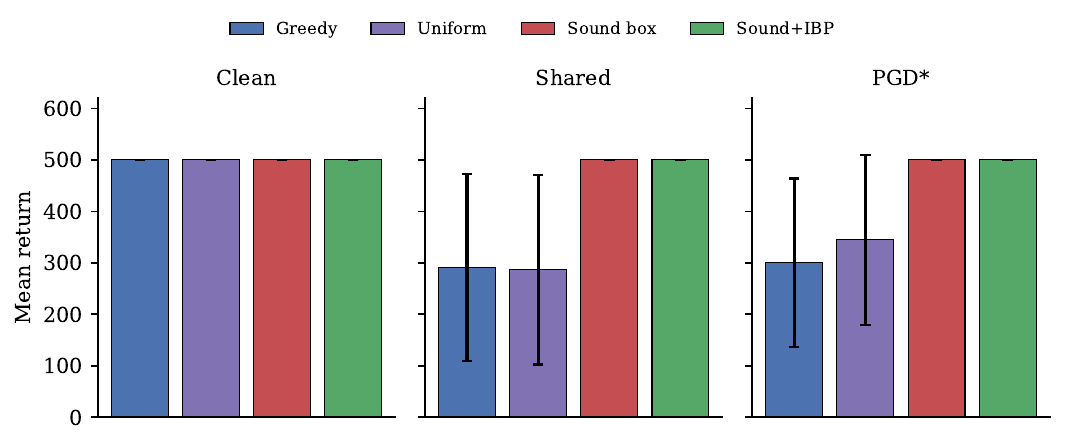}
\caption{Frozen vanilla CartPole $Q$, the four test-time rules of Table~\ref{tab:cp-sound}, $\varepsilon{=}0.05$, five seeds.
Shared-random and PGD* drop greedy and uniform-sample maximin.
The sound box and the interval-min variant of that box hold $500$.
Matched SB3 is omitted.}
\label{fig:cp}
\end{figure}

The main CartPole comparison is the trained sound box in \S\ref{sec:cp}.
This appendix keeps the frozen-weight rule swap, the failed GRU-particle port, pole-sensor attacks, the $\varepsilon$ sweep, and a position-sensor variant.
Figure~\ref{fig:cp} plots the four frozen rules in Table~\ref{tab:cp-sound}.

\paragraph{Frozen vanilla $Q$.}
Table~\ref{tab:cp-sound} and Figure~\ref{fig:cp} put four test-time rules on the untouched per-seed vanilla checkpoint ($\varepsilon{=}0.05$, $50$ episodes, five seeds).
All four solve clean CartPole.
Shared-random and PGD* drop greedy to $290.9{\pm}181.3$ and $300.1{\pm}163.8$.
The sound box and the same box with an interval inner min both hold $500.0{\pm}0.0$ on those two columns.
Uniform-sample maximin does not ($286.5{\pm}184.2$ shared, $345.0{\pm}165.3$ PGD*) and is worse than greedy against a matched SB3 attacker ($78.7{\pm}72.9$ against $212.6{\pm}262.5$, with no vacuous seeds).
The sound matched column is vacuous on all five seeds and is omitted.
The hold at $500$ is the test-time rule on weights that were not fine-tuned.

\begin{table*}[t]
\centering
\small
\caption{Four decision rules on frozen vanilla CartPole weights, $\varepsilon{=}0.05$, $50$ episodes, five seeds.
Mean $\pm$ sample std of the per-seed means.
The matched column is omitted for the sound rows because the SB3 attacker never ended an episode on any seed.
Greedy matched is vacuous on seed~$2$ only.}
\label{tab:cp-sound}
\setlength{\tabcolsep}{5pt}
\begin{tabular}{@{}lcccc@{}}
\toprule
Rule & Clean & Shared-random & PGD* & Matched SB3 \\
\midrule
Greedy & $500.0{\pm}0.0$ & $290.9{\pm}181.3$ & $300.1{\pm}163.8$ & $212.6{\pm}262.5$ \\
Uniform sample & $500.0{\pm}0.0$ & $286.5{\pm}184.2$ & $345.0{\pm}165.3$ & $78.7{\pm}72.9$ \\
Sound box & $500.0{\pm}0.0$ & $500.0{\pm}0.0$ & $500.0{\pm}0.0$ & --- \\
Sound + interval min & $500.0{\pm}0.0$ & $500.0{\pm}0.0$ & $500.0{\pm}0.0$ & --- \\
\bottomrule
\end{tabular}
\end{table*}

\begin{figure}[t]
\centering
\includegraphics[width=\columnwidth]{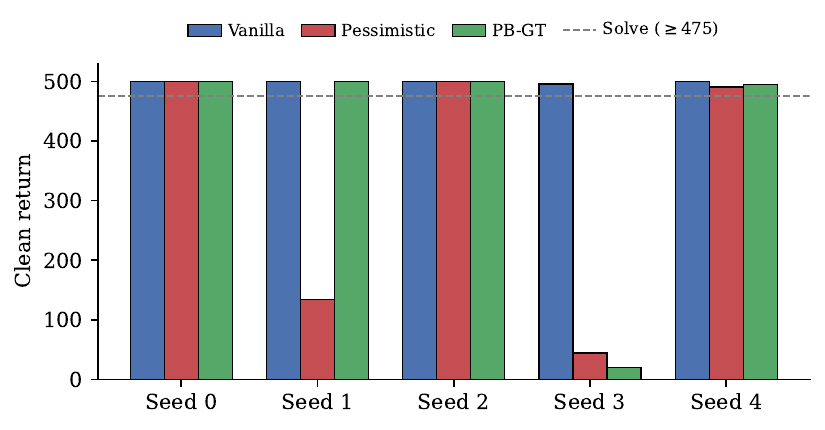}
\caption{Per-seed clean return after the GRU-particle fine-tune, greedy readout.
The main CartPole experiment uses the sound box instead.}
\label{fig:cp-seeds}
\end{figure}

\paragraph{Failed GRU-particle port.}
A CartPole port used Double DQN with a GRU particle belief, trained for five seeds with the vanilla checkpoint restored, and evaluated for $100$ episodes.
Table~\ref{tab:cp-gru} and Figure~\ref{fig:cp-seeds} record that port.
Vanilla's clean return is $499.1{\pm}1.9$.
If the pessimistic fine-tune is scored with the maximin rule, clean return is $132.1{\pm}126.1$.
A greedy readout of those same weights is $333.6{\pm}225.8$, and PB-GT's greedy clean return is $403.1{\pm}214.1$.
Matched-attacker return for the pessimistic port is $11.0{\pm}2.9$ on every seed in the table of means.
Matched-attacker inference in that port was $1.88$\,ms/step for the particle maximin and $0.037$\,ms/step for the greedy test.
The main result uses the sound box of Table~\ref{tab:cp} instead.

\begin{table}[t]
\centering
\small
\caption{GRU-particle CartPole port, five seeds, $100$ episodes.
Mean $\pm$ sample std of the per-seed means.
Pessimistic clean is the maximin readout.
This table is not the main CartPole result.}
\label{tab:cp-gru}
\setlength{\tabcolsep}{3pt}
\begin{tabular}{@{}lccc@{}}
\toprule
Metric & Vanilla & Pes.\ (GRU) & PB-GT \\
\midrule
Clean & $499.1{\pm}1.9$ & $132.1{\pm}126.1$ & $403.1{\pm}214.1$ \\
Matched attacker & $113.4{\pm}216.3$ & $11.0{\pm}2.9$ & $108.9{\pm}218.7$ \\
Shared-random & $284.0{\pm}175.9$ & $124.6{\pm}185.2$ & $149.5{\pm}208.5$ \\
PGD* & $318.8{\pm}154.2$ & $131.0{\pm}126.0$ & $189.0{\pm}147.9$ \\
\bottomrule
\end{tabular}
\end{table}

\paragraph{Pole-sensor attacks.}
Table~\ref{tab:cp-bias} adds two structured pole-sensor attacks at $\varepsilon{=}0.05$ (five seeds, $20$ episodes) on frozen vanilla $Q$.
Worst-of-$9$ constant bias drops greedy to $178.8{\pm}189.7$.
The sound box is the highest mean of the three rules ($428.6{\pm}146.6$) and is not $500$ on every seed.
Exhaustive one-step search is weaker against greedy ($403.1{\pm}201.8$) and lowers uniform ($252.4{\pm}226.6$).
The sound mean on that attack is $419.3{\pm}180.4$.

\begin{table}[t]
\centering
\small
\caption{Frozen vanilla CartPole $Q$, pole-sensor attacks, $\varepsilon{=}0.05$, five seeds, $20$ episodes.
Worst-of-$9$ is the min over the nine constant biases in $\{-\varepsilon,0,\varepsilon\}^2$ on $(\theta,\dot\theta)$.
Exhaustive search picks the bias with the lowest $Q(s,\pi(\tilde s))$ at each step.}
\label{tab:cp-bias}
\setlength{\tabcolsep}{4pt}
\begin{tabular}{@{}lccc@{}}
\toprule
Attack & Greedy & Uniform & Sound \\
\midrule
Worst-of-$9$ & $178.8{\pm}189.7$ & $312.1{\pm}208.7$ & $428.6{\pm}146.6$ \\
Exhaustive $1$-step & $403.1{\pm}201.8$ & $252.4{\pm}226.6$ & $419.3{\pm}180.4$ \\
\bottomrule
\end{tabular}
\end{table}

\paragraph{Pole-sensor $\varepsilon$ sweep.}
Table~\ref{tab:cp-eps} averages two seeds and $20$ episodes.
Attacks perturb only $(\theta,\dot\theta)$.
The sound box holds $500$ at $\varepsilon\in\{0.025,0.05\}$ on those two seeds and drops at $\varepsilon{=}0.10$: worst-of-$9$ sound $139.9$, greedy $10.6$, uniform $74.0$.
The five-seed check at $\varepsilon{=}0.05$ is Table~\ref{tab:cp-bias}.
At $\varepsilon\in\{0.15,0.20\}$ the worst-of-$9$ sound mean is $9.1$.

\begin{table}[h]
\centering
\small
\caption{CartPole pole-sensor $\varepsilon$ sweep on frozen vanilla $Q$.
Mean return over two seeds and $20$ episodes.
Attack definitions match Table~\ref{tab:cp-bias}.}
\label{tab:cp-eps}
\setlength{\tabcolsep}{3.5pt}
\begin{tabular}{@{}llccc@{}}
\toprule
$\varepsilon$ & Attack & Greedy & Uniform & Sound \\
\midrule
$0.025$ & worst-of-$9$ & $329.8$ & $487.8$ & $500.0$ \\
$0.025$ & exhaustive & $141.2$ & $470.1$ & $500.0$ \\
$0.05$ & worst-of-$9$ & $51.0$ & $439.3$ & $500.0$ \\
$0.05$ & exhaustive & $488.0$ & $462.0$ & $500.0$ \\
$0.10$ & worst-of-$9$ & $10.6$ & $74.0$ & $139.9$ \\
$0.10$ & exhaustive & $500.0$ & $133.2$ & $190.6$ \\
$0.15$ & worst-of-$9$ & $16.8$ & $14.8$ & $9.1$ \\
$0.15$ & exhaustive & $371.2$ & $13.8$ & $86.6$ \\
$0.20$ & worst-of-$9$ & $33.0$ & $9.1$ & $9.1$ \\
$0.20$ & exhaustive & $14.8$ & $14.9$ & $29.9$ \\
\bottomrule
\end{tabular}
\end{table}

\paragraph{Position-sensor restriction.}
When only $(x,\theta)$ are attacked and the velocities are observed cleanly, with $\varepsilon_{\mathrm{vec}}=[0.05,0,0.05,0]$, the intersection can contract.
On the sound-box log the filtered widths average $0.0089$ in $x$ and $0.0089$ in $\theta$, against an unfiltered width of $2\varepsilon=0.10$, with velocity widths about $0.0002$ and no empty intersections across the five seeds.
Shared-random and PGD* still hold the sound box at $500$, against greedy $276.6{\pm}201.5$ and $290.3{\pm}170.4$ (uniform $290.9{\pm}178.7$ and $204.6{\pm}176.7$).
The sound matched mean is $494.3{\pm}12.7$ and is vacuous on $4$ of $5$ seeds.
We do not treat that column as a robustness claim.

\begin{table*}[t]
\centering
\small
\caption{Frozen vanilla CartPole $Q$, position-sensor threat $\varepsilon_{\mathrm{vec}}=[0.05,0,0.05,0]$, $50$ episodes, five seeds.
The sound matched column is vacuous on $4$ of $5$ seeds.}
\label{tab:cp-pos}
\setlength{\tabcolsep}{5pt}
\begin{tabular}{@{}lcccc@{}}
\toprule
Rule & Clean & Shared-random & PGD* & Matched SB3 \\
\midrule
Greedy & $500.0{\pm}0.0$ & $276.6{\pm}201.5$ & $290.3{\pm}170.4$ & $134.7{\pm}205.3$ \\
Uniform sample & $498.1{\pm}4.3$ & $290.9{\pm}178.7$ & $204.6{\pm}176.7$ & $139.5{\pm}206.1$ \\
Sound box & $500.0{\pm}0.0$ & $500.0{\pm}0.0$ & $500.0{\pm}0.0$ & $494.3{\pm}12.7$ \\
Sound + interval min & $500.0{\pm}0.0$ & $500.0{\pm}0.0$ & $500.0{\pm}0.0$ & $494.3{\pm}12.7$ \\
\bottomrule
\end{tabular}
\end{table*}

\end{document}